\documentclass[lettersize,journal]{IEEEtran}
\usepackage{amsmath,amsfonts}
\usepackage{algorithmic}
\usepackage{algorithm}
\usepackage{array}
\usepackage[caption=false,font=normalsize,labelfont=sf,textfont=sf]{subfig}
\usepackage{textcomp}
\usepackage{stfloats}
\usepackage{url}
\usepackage{verbatim}
\usepackage{graphicx}
\usepackage{cite}

\usepackage{rotating}
\usepackage{makecell}
\usepackage{tabularray}
\usepackage{xtab,afterpage}
\usepackage{xltabular}
\usepackage{threeparttablex}
\usepackage{booktabs,caption}
\usepackage{pdflscape}
\usepackage[dvipsnames]{xcolor}

\makeatletter
 \let\old@ps@headings\ps@headings
 \let\old@ps@IEEEtitlepagestyle\ps@IEEEtitlepagestyle
 \def\confheader#1{%
 \def\ps@headings{%
 \old@ps@headings%
 \def\@oddhead{\strut\hfill#1\hfill\strut}%
 \def\@evenhead{\strut\hfill#1\hfill\strut}%
 }%
 \def\ps@IEEEtitlepagestyle{%
 \old@ps@IEEEtitlepagestyle%
 \def\@oddhead{\strut\hfill#1\hfill\strut}%
 \def\@evenhead{\strut\hfill#1\hfill\strut}%
 }%
 \ps@headings%
 }
 \makeatother

 \usepackage[pscoord]{eso-pic}
\newcommand{\placetextbox}[3]{
 \setbox0=\hbox{#3}
 \AddToShipoutPictureFG*{ \put(\LenToUnit{#1\paperwidth},\LenToUnit{#2\paperheight}){\vtop{{\null}\makebox[0pt][c]{#3}}}
 }
 }

  \placetextbox{.5}{0.055}{\footnotesize{© 2023 IEEE. Personal use of this material is permitted. Permission from IEEE must be obtained for all other uses, in any current or future media,}} 
  \placetextbox{.5}{0.045}{\footnotesize{including reprinting/republishing this material for advertising or promotional purposes, creating new collective works, for resale or redistribution}}
  \placetextbox{.5}{0.035}{\footnotesize{to servers or lists, or reuse of any copyrighted component of this work in other works.}}

\begin{document}

\title{Pedestrian Trajectory Prediction in Pedestrian-Vehicle Mixed Environments: A Systematic Review}

\author{Mahsa Golchoubian$^{1\dag}$, Moojan Ghafurian$^{2}$, Kerstin Dautenhahn$^{2}$,  Nasser Lashgarian Azad$^{1}$
\thanks{\dag Corresponding Author: Mahsa Golchoubian {\tt\small mahsa.golchoubian@uwaterloo.ca}}
\thanks{$^{1}$Department of Systems Design Engineering, University of Waterloo, Canada}
\thanks{$^{2}$Department of Electrical and Computer Engineering, University of Waterloo, Canada}%
}



\maketitle

\begin{abstract}
Planning an autonomous vehicle's (AV) path in a space shared with pedestrians requires reasoning about pedestrians’ future trajectories. A practical pedestrian trajectory prediction algorithm for the use of AVs needs to consider the effect of the vehicle’s interactions with the pedestrians on pedestrians' future motion behaviours. In this regard, this paper systematically reviews different methods proposed in the literature for modelling pedestrian trajectory prediction in presence of vehicles that can be applied for unstructured environments. This paper also investigates specific considerations for pedestrian-vehicle interaction (compared with pedestrian-pedestrian interaction) and reviews how different variables such as prediction uncertainties and behavioural differences are accounted for in the previously proposed prediction models. PRISMA guidelines were followed. Articles that did not consider vehicle and pedestrian interactions or actual trajectories, and articles that only focused on road crossing were excluded. A total of 1260 unique peer-reviewed articles from ACM Digital Library, IEEE Xplore, and Scopus databases were identified in the search. 64 articles were included in the final review as they met the inclusion and exclusion criteria. An overview of datasets containing trajectory data of both pedestrians and vehicles used by the reviewed papers has been provided. Research gaps and directions for future work, such as having more effective definition of interacting agents in deep learning methods and the need for gathering more datasets of mixed traffic in unstructured environments are discussed.  
\end{abstract}


\section{Introduction}


With the growing number of research on autonomous vehicles (AV) \cite{fan2019key}, we will eventually have these vehicles operating close to pedestrians in crowded more unstructured environments in the future. Therefore, there is a need for understanding how these AVs are going to deal with crowds of pedestrians beyond the framework forced by the traffic laws on the roads and in everyday encounters governed more by social etiquette.

Modelling pedestrians' motion behaviours for predicting their future trajectory is a crucial skill for autonomous vehicles that want to navigate in an environment shared with pedestrians. Pedestrians' future motions are affected by many factors, such as the surrounding agents (e.g., pedestrians and vehicles) known as interaction effects. Therefore, safe operation of an autonomous vehicle relies on modelling the impact of the vehicle's presence and actions on pedestrians' future trajectories. 

Many pedestrian trajectory prediction models have been proposed that include only pedestrian-pedestrian interaction without considering any vehicle in their model. These models can hardly be used for autonomous vehicle (AV) algorithms as they do not capture the effect of the vehicle's motion on the pedestrian's decisions and motions. Among the papers that include the pedestrian-vehicle interaction in the pedestrian's motion model, a great focus has been put on pedestrian road crossing behaviour \cite{rasouli2019autonomous,zhang2015review} that takes into account the strict structure of the conventional roads while less paper can be found studying pedestrian-vehicle interaction in off-road, unstructured environments.

However, modelling interaction effects (between vehicles and pedestrians) becomes more complicated in environments with no pre-specified paths (e.g., lanes, crosswalks) or strict traffic rules, known as unstructured environments. These unstructured environments are defined qualitatively as spaces where no clear lane marks or road dividers exist and no strict traffic rules are followed by the traffic agents ~\cite{jyothi2019driver,kerscher2018intention}. Quantitative measures for distinguishing unstructured environments from structured spaces based on motion features are proposed in \cite{Golchoubian2022} and it was shown that pedestrian motion behaviours are different in a structured versus unstructured environment \cite{Golchoubian2022}. While currently, the main focus of many researchers and industries is on designing autonomous vehicles (AV) that operate in the structure of the roads, low-speed AVs will eventually enter into more unstructured environments for which they need to handle more diverse interaction scenarios with the pedestrians that are sharing the same space. Examples of this application domain can be future low-speed autonomous vehicles that will be used as a mobility aid for transporting passengers around in off-road outdoor (e.g., parks, campus environments) and indoor environments (e.g., shopping malls, airport terminals).

Therefore, the focus of this review paper will be on pedestrian trajectory prediction literature with the two key factors of 1) modelling pedestrian-vehicle interaction and 2) studying these interactions and trajectories in models that can be applied to unstructured environments. 

Pedestrian motion modelling is the process of finding the formulation for the pedestrian's transition from one state to the next state over time. This transition gets influenced by the pedestrian's previous states and the presence of nearby agents. The later effect is usually captured, to some extent, as the interaction effect part of the proposed motion models. Therefore, in this review, when referring to motion models we mean a model that also embeds interaction effects. These models can then be used for generating pedestrians' trajectories over time in a simulation or predicting the pedestrians' next state given a history of their trajectory and information from their surrounding neighbours.
Having said that, the process of finding a motion model that can describe all the influential factors on a pedestrian's state transition is almost the same as discovering a trajectory prediction model. Therefore, in this paper both \textit{trajectory prediction models} and \textit{motion models} in the literature are studied and we will refer to them as \textit{motion prediction modelling approaches} in general.

There already exists a couple of survey papers on pedestrian trajectory prediction. Models for both pedestrian and vehicle trajectory prediction are reviewed in \cite{gulzar2021survey,kolekar2021behavior} while showing the similarities between the motion models used for vehicles and pedestrians \cite{gulzar2021survey}. But the authors have not discussed the differences between the various agent classes and their interactions caused by their different dynamic nature which is one of the focuses of our work. Kolekar et al. \cite{kolekar2021behavior} pointed out the importance of using the traffic rules and road geometry as input for a promising behaviour prediction model since they were focused on road scenarios, whereas our work is focused on reviewing the existing models that can be applied to unstructured environments (i.e., those without assumptions on traffic rules). In addition, not all the papers surveyed above necessarily have both types of agents simultaneously to consider pedestrian-vehicle interaction. 

Some papers have restricted their survey to studies proposing only deep learning models \cite{sighencea2021review,chen2021survey}, showing the shift of the recent models toward network architectures that can capture interaction affects \cite{chen2021survey}. This is while others have covered a larger variety of methods, including both empirical science and machine learning methods~\cite{rudenko2020human,camara2020pedestrian} which is also an approach followed in this review. Rudenko et al \cite{rudenko2020human}, have categorized proposed human trajectory prediction models based on their motion modelling approaches and the contextual information used in the models. The interaction between the agents is discussed as part of the contextual cues of a dynamic environment considered for trajectory prediction. While summarizing the strengths and weaknesses of different modelling approaches  \cite{rudenko2020human}, they have mentioned how nearby agents' dynamic cues can be handled within each model category. But unlike the current review, they do not distinguish between pedestrian-vehicle interaction compared to pedestrian-pedestrian interaction.
Gulzar et al. (2021)~\cite{gulzar2021survey} discuss interactions between agents as part of the situational awareness of the agent's model. They suggest that models relying on data can capture more complex interactions between agents compared to physics-based models. Pedestrian interaction models are also discussed in \cite{camara2020pedestrian} while devoting a separate section to the game theoretic approaches taken for modelling mutual interactions and pointing out that more research is needed for bringing psychological-informed models of human interaction into motion behaviour prediction models.

However, not all the papers surveyed above necessarily have both pedestrians and vehicles simultaneously in the environment. Therefore, the interaction methods surveyed in the above papers mainly include only pedestrian-pedestrian interactions with few also covering the interactions between a pedestrian and a vehicle during a road crossing. Hence, to the best of our knowledge, this is the first comprehensive systematic review of the approaches taken for modelling pedestrian-vehicle interaction in pedestrian motion prediction methods, focusing on models applicable to unstructured environments. Within this review, special attention is paid to how pedestrian-vehicle interaction is encoded differently from pedestrian-pedestrian interaction in an environment with heterogeneous agent types as these interactions are different in nature due to the size, speed and motion kinematics differences of various agent types. In this regard, the datasets that include different agent types and were used for pedestrian trajectory prediction in the presence of a vehicle are also summarized as part of the survey.

\begin{figure}[t!]
  \centering
  \includegraphics[width=1\linewidth]{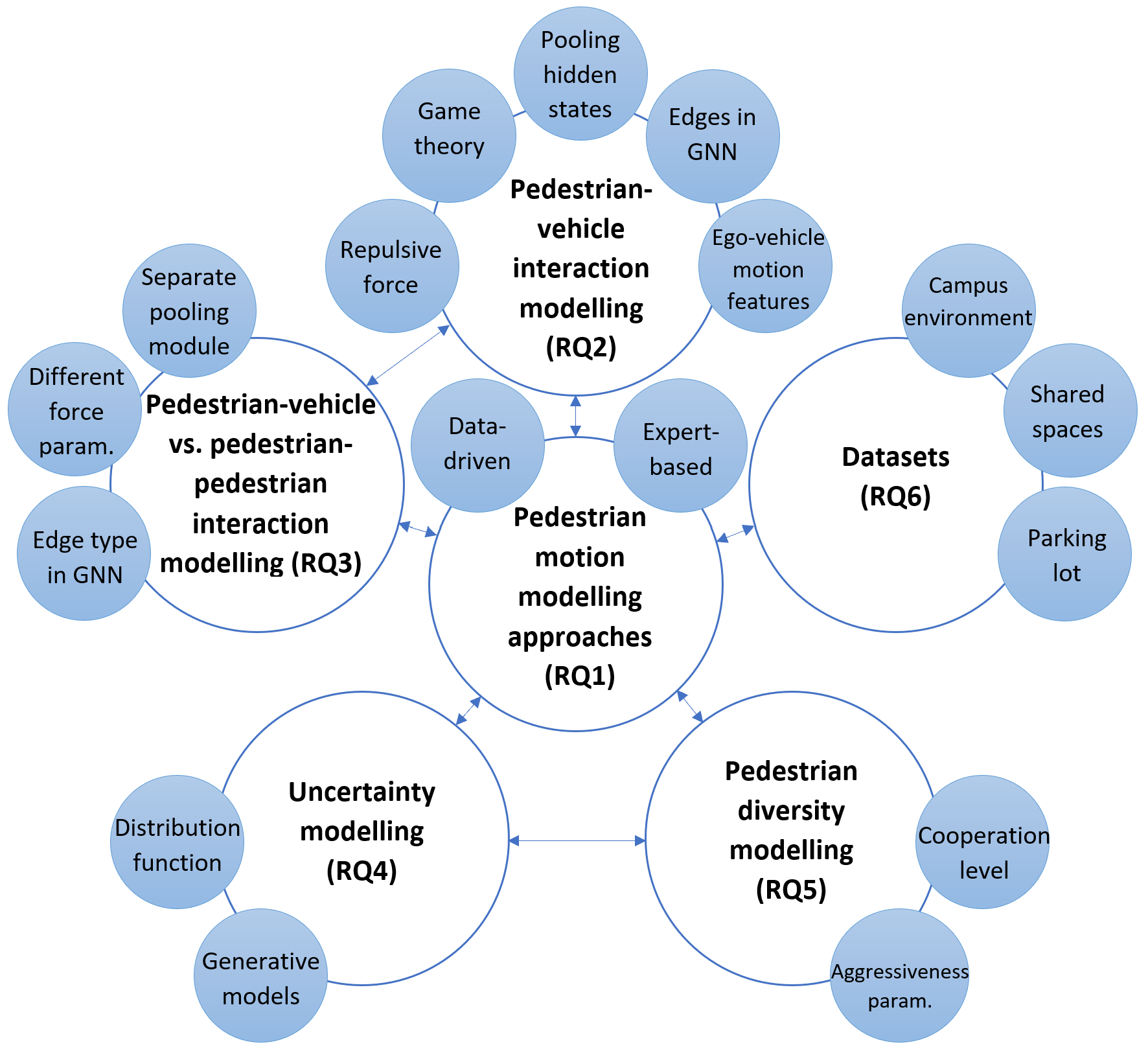}
  \caption{Main topics covered in the review through answering the six research questions. While many of these topics can be interconnected, only the main connections between our research questions are visualized. Examples of the categories found in each topic are shown in the smaller circles.}
  \label{fig:Overv}
\end{figure}

It should also be kept in mind that pedestrians can handle the same interaction scenario differently due to individual differences and other factors affecting people's actions, such as their different internal states, emotional attitudes, time pressure, or perception of the environment. Therefore, we also review the methods proposed in the literature for considering these pedestrian differences as well as the approaches taken for modelling the overall stochastic nature of the pedestrian's trajectory prediction. The details of our research questions are listed below (also see Fig. \ref{fig:Overv}).

\textbf{Research questions:}
\begin{itemize}
    \item \textbf{RQ1}: What are the existing methods for pedestrian trajectory prediction and motion modelling under the influence of vehicles, which can be applied to unstructured environments?

    \item \textbf{RQ2}: How are interactions between pedestrians and vehicles modelled when predicting the pedestrian’s trajectory?

    \item \textbf{RQ3}: How are the differences between multiple traffic modes (e.g., pedestrian, vehicle) and multiple types of interactions accounted for in these past models?

    \item \textbf{RQ4}: If considered, how is uncertainty of pedestrians' future position modelled in these past models?

    \item \textbf{RQ5}: If considered, how is diversity in pedestrian behaviour encoded in these models? (hidden characteristics of personality, gender, age, level of cooperation etc.)

    \item \textbf{RQ6}: Which datasets are collected or have been used in the previous models and what type of data do they contain?
\end{itemize}

\section{Review Method}

For answering our research questions, we conducted a systematic review by following the PRISMA \footnote{Preferred Reporting Items for Systematic Reviews and Meta-Analyses (PRISMA)} guidelines. The search query used for collecting papers on pedestrian trajectory prediction or motion modelling in mixed pedestrian-vehicle environments is given in Table \ref{tab:Search}. We came up with this search query through an iterative process in Scopus in consultation with University of Waterloo's Librarian. We searched for these terms in the title, abstract and keywords of papers.

\begin{table*}
\centering
\footnotesize
\caption{The two search queries used in Scopus. These two queries were combined together with OR} \label{tab:Search}
{\begin{tabular}{|p{1.2cm}|p{16cm}|}
\hline
& ((trajector* OR motion OR path) W/5 (pedestrian OR pedestrians OR human OR crowd OR "mixed traffic" OR "Heterogeneous traffic")) OR ((pedestrian OR pedestrians) W/5 (behavior OR behaviour)) \\
& \textbf{AND} \\
\textbf{Trajectory} & (predict* OR forecast*) W/10 (trajector* OR motion* OR path OR behavior OR behaviour) \\
\textbf{prediction} & \textbf{AND} \\
& vehicle OR vehicles OR car OR cars OR "shared space" OR "shared spaces" OR ((mixed OR heterogeneous OR multimodal OR multi-modal) PRE/1 (traffic OR transport* OR "road user" OR "road users" OR agent)) \\
\hline
& (model* W/10 (trajector* OR motion OR behavior OR behaviour OR interaction)) W/10 (pedestrian OR pedestrians OR "mixed traffic" OR "shared space" OR "shared spaces") \\
\textbf{Motion} & \textbf{AND} \\
\textbf{modelling} & vehicle OR vehicles OR car OR cars OR "shared space" OR "shared spaces" OR ((mixed OR heterogeneous OR multimodal OR multi-modal) PRE/1 (traffic OR transport* OR "road user" OR "road users" OR agent)) \\
\hline
\end{tabular}}
\end{table*}

We used the three databases of Scopus, IEEE Xplore, ACM DL for conducting our search and collecting the related papers. The search query in Table \ref{tab:Search} was adjusted for each database due to their different styles and requirements (e.g., using fewer terms in IEEE and ACM due to the word limit of their search engine). We limited our search to peer-reviewed articles published in journals or conference proceedings within the past 10 years (2012-2022). We conducted the search on June 2022, which identified a total of 933 papers in Scopus, 524 in IEEE Xplore, and 113 documents in ACM DL. After removing the duplicates we were left with 1260 papers to screen. 

The inclusion and exclusion criteria used for extracting the papers are listed as follows.

\textbf{Inclusion Criteria}:

\begin{itemize}
    \item Articles on pedestrian trajectory prediction or motion modelling
	\item Articles considering the presence of both pedestrians and vehicles in the environment (mixed traffic)
    \item Articles that modelled the interactions between the agents (pedestrians and vehicles)
	\item Articles that involved unstructured or shared spaces (as defined above)
    \item Articles with no restriction on the interaction scenarios and no use of traffic rules or road-specific information (e.g, road boundaries and sizes) in the model
    \item Peer-reviewed articles published in journals or conference proceedings
    \item Articles published in the past 10 years (2012-2022)
\end{itemize}

\textbf{Exclusion Criteria}:

\begin{itemize}
    \item Articles considering  only vehicles or only pedestrian trajectories (focusing on homogeneous user types)
    \item Articles that did not consider the interactions between the agents
    \item Articles that exclusively focused on road crossing
    \item Articles that investigated pedestrian trajectory prediction for small non-vehicular robots in an unstructured environment
    \item Articles that did not focus on actual trajectories and only predicted intentions or decisions
    \item Articles that did not model the trajectories at the individual level and focused on the flow of pedestrians (e.g. evacuation models) 
    \item Articles not written in English
\end{itemize}

\begin{figure}[t!]
  \centering
  \includegraphics[width=1\linewidth]{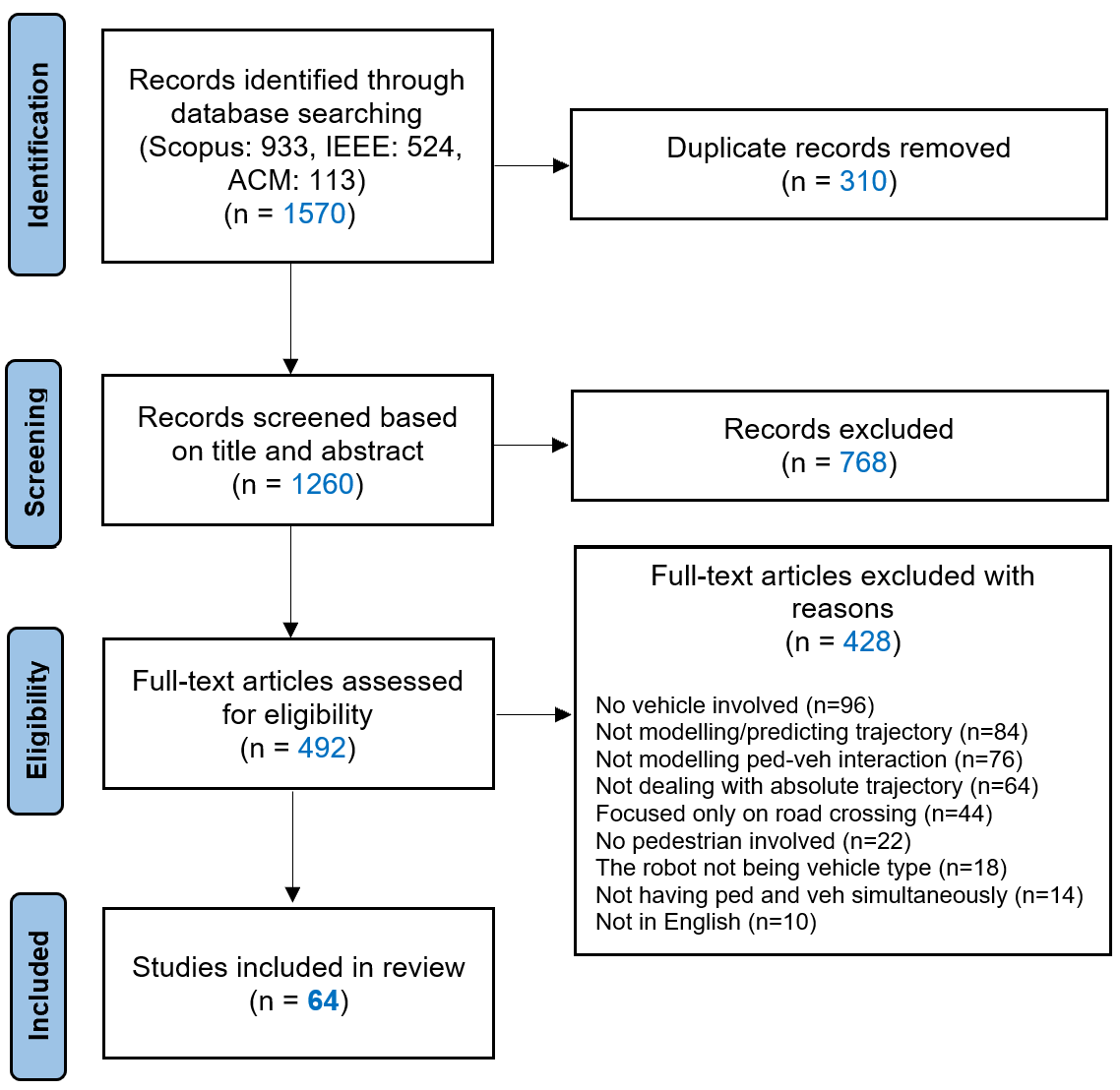}
  \caption{PRISMA guidance flow diagram}
  \label{fig:PRISMA}
\end{figure}

After title and abstract screening (performed by one of the authors), a total of 492 articles remained, the full text of which were assessed for Eligibility. Fig. \ref{fig:PRISMA} shows the PRISMA flow diagram. A final set of 64 papers met all inclusion and exclusion criteria and were left for a full review.

In order to find models that can be used for unstructured environments, we looked for papers that were modelling pedestrian's trajectory while interacting with vehicles in (1) off-road environments, (2) environments with alleviated structure and traffic rules, or (3) urban street environments were the model did not use road-specific information and did not restrict the pedestrian-vehicle interaction scenario to road crossing. One example of environments with alleviated structure and traffic rules is the shared spaces which are urban environments with removed curbs, road surface markings, and traffic signs. Different traffic agents in these environments are encouraged to share the space and negotiate the priority based on social rules~\cite{predhumeau2021pedestrian}. Therefore, diverse interaction scenarios with different approach directions similar to off-road interaction can happen in these environments. A couple of trajectory dataset from such environments are available that were used in the literature for training or calibrating the parameters of prediction models. On the other hand, we excluded papers that were focusing exclusively on pedestrian crossing scenarios happening on conventional roads as they were clearly happening in a structured environment and using traffic rules or road dimensions. But we kept papers that, despite using datasets of urban street environments, did not restrict the pedestrian-vehicle interaction scenario in their model to only crossing and did not use road-specific information. This means that these models can also be used for unstructured environments. Therefore, we included them to have a more comprehensive review of pedestrian-vehicle interaction modelling in proposed pedestrian trajectory prediction methods.

In the rest of the paper, sections \ref{Q1}-\ref{Q6} discuss results related to each of the research questions. In section \ref{Future}, we discuss the identified gaps and provide suggestions about the future research directions. We conclude the paper with section \ref{Conc}.

\section{RQ1: Pedestrian Motion Prediction Modeling Approaches} \label{Q1}

\begin{figure}[t!]
  \centering
  \includegraphics[width=0.7\linewidth]{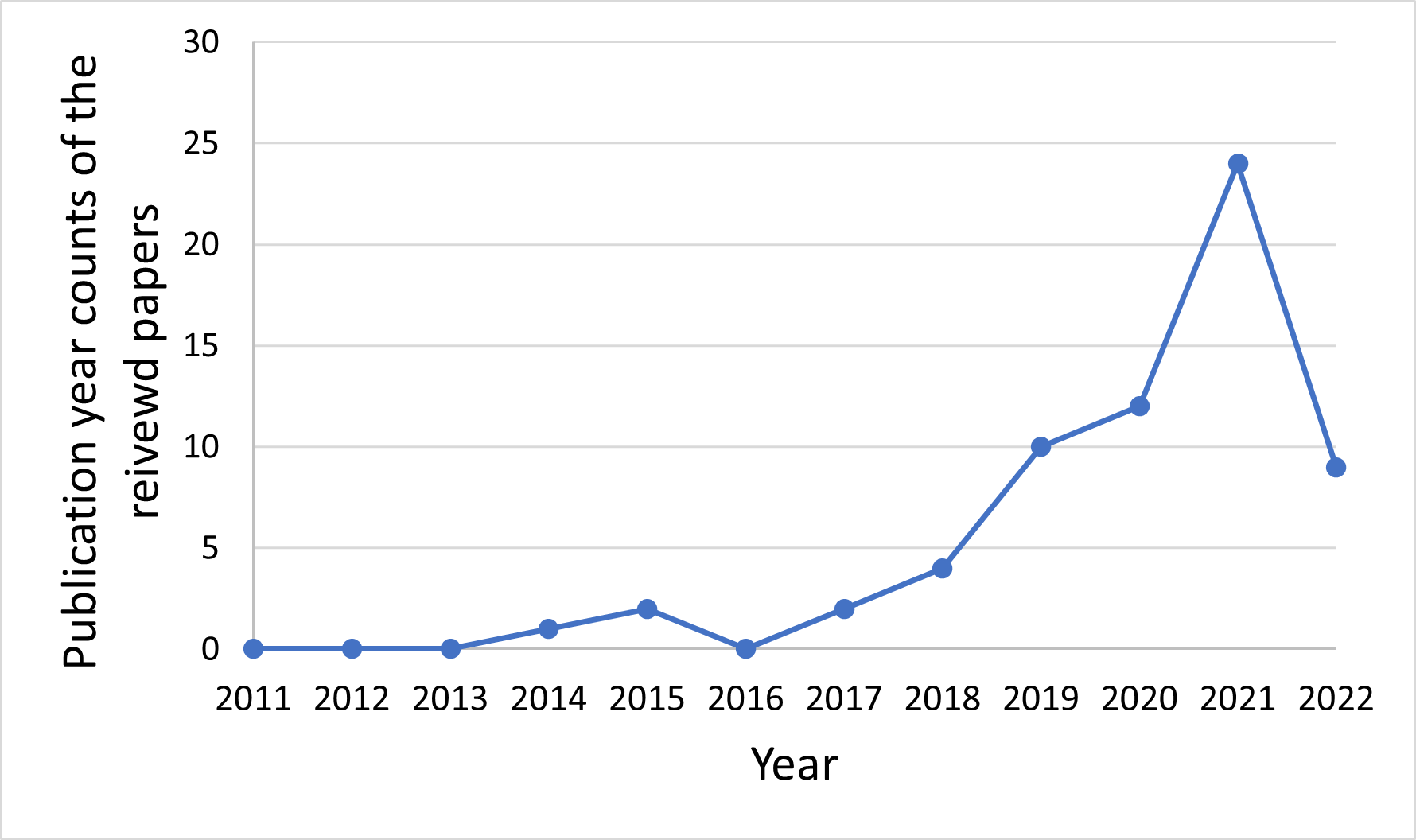}
  \caption{Distribution of publication year of the reviewed paper between 2011 until June 2022}
  \label{fig:Year}
\end{figure}

The year distribution of the 64 extracted articles is shown in Fig \ref{fig:Year}, depicting the fact that more focus is being put in recent years on trajectory prediction of heterogeneous agents including the pedestrian-vehicle interaction modelling.

Different methods have been used in the reviewed literature for modelling and predicting pedestrian trajectories over time. 
The reasoning mechanism behind how a pedestrian’s state evolves over time is either formulated through explicit hand-crafted rules or learned from real-world trajectory data. Therefore, the two main categories found in the literature for pedestrian motion prediction are expert-based and data-driven approaches while in some proposed methods a combination of both is used.

Due to the focus of this survey on papers that include interactions, all the approaches reviewed here include interaction modelling within their motion model. Therefore, in this section, an overview of the interaction-aware prediction approaches pursued in the literature is provided without distinguishing their interaction modelling part from the rest of the model. However, a more detailed discussion on the specific interaction modelling part will be left to section \ref{Q2}.

\begin{figure}[b!]
  \centering
  \includegraphics[width=0.95\linewidth]{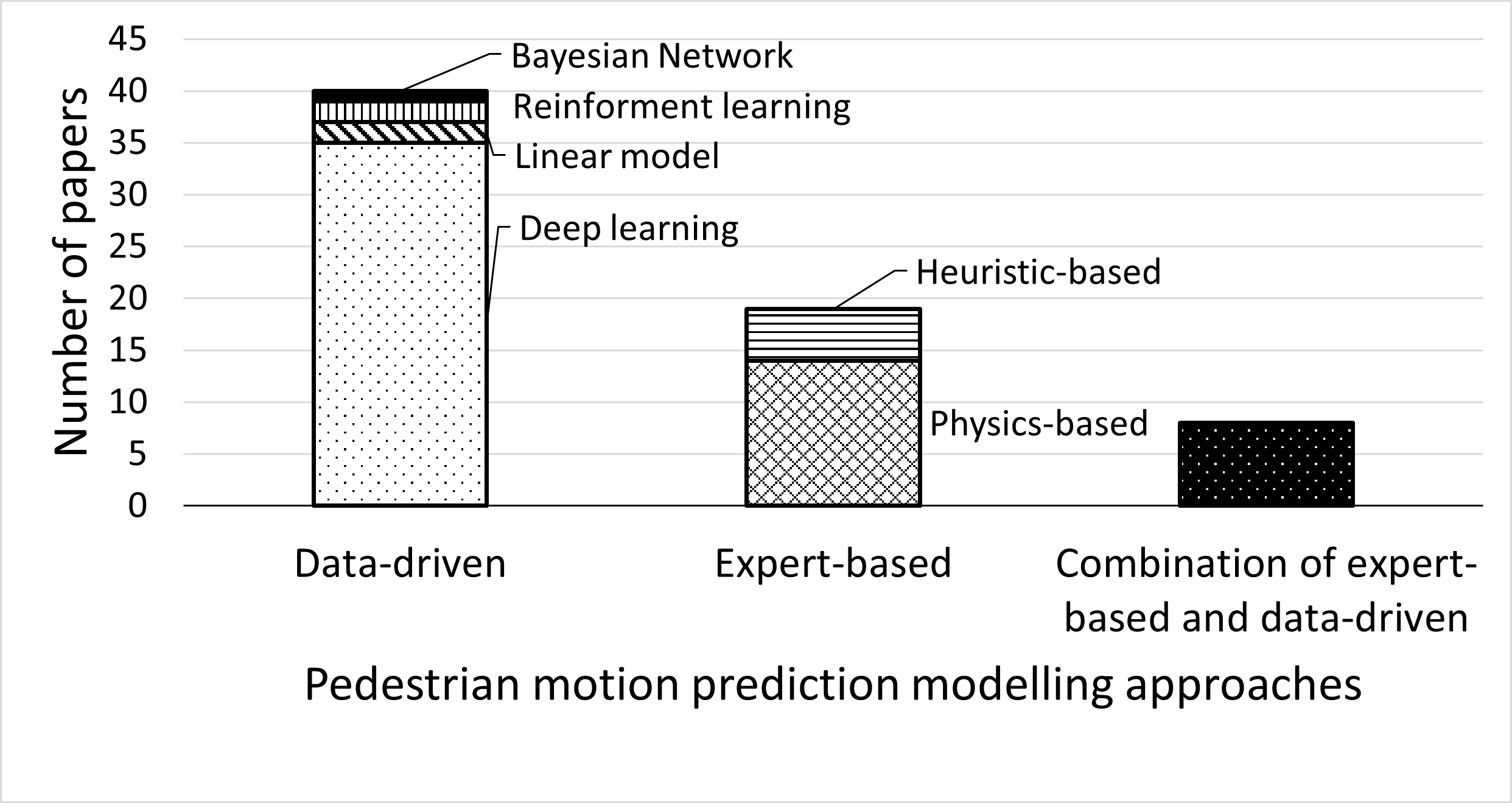}
  \caption{The different pedestrian motion prediction approaches used in the literature with their identified sub-categories. The number of reviewed papers in each category is shown.}
  \label{fig:Motion}
\end{figure}

Motion prediction approaches in Table~\ref{tab:summary} summarizes the model used in each of the reviewed articles. Fig. \ref{fig:Motion} shows the number of reviewed papers that uses each of these approaches and its sub-categories. We came up with these categories based on the papers that we reviewed while being inspired by the existing categories in the literature \cite{rudenko2020human, camara2020pedestrian,cheng2021trajectory,gulzar2021survey}. 

\subsection{Expert-based Models}
Expert-based models involve a human-crafted explicit rule that governs pedestrians' state transition or represents their decision-making process. 
Physics-based models governed by laws of physics and other heuristic-based transition models in discrete time fall under this category.

\subsubsection{\textbf{Physics-based models}}
Physics-based models use equations of motion from laws of physics for propagating forward the state of the agent. These models use either kinematics or dynamics equations derived from laws of physics (e.g., Newton's second law)\cite{rudenko2020human}. The Social Force Model (SFM) as a physics-based model uses the second law of motion for simulating pedestrians' motion toward their goal position.

In the dynamic equation of SFM, the motion of the pedestrian is formulated to be under the influence of multiple forces with the basic ones being an attractive force to the destination and repulsive forces from both obstacles and other agents (e.g, pedestrians or vehicles) in the environment. The summation of these forces specifies the acceleration vector of a pedestrian according to Newton's second law of motion \cite{yang2018crowd,rinke2017multi}. 
While the underlying dynamic equation of SFM with only the attraction force towards the destination can be used as a basic motion model for pedestrians, the social forces from other agents are an additional factor in the SFM for interaction modelling. Therefore, some papers have used the SFM as an interaction model in a multi-layer architecture on top of a simple motion planner that calculates the agent's shortest path to its destination considering only the static obstacles in the environment ~\cite{anvari2015modelling,johora2018modeling,johora2021transferability}.

Other physics-based models based on kinematic equations are the constant velocity and constant acceleration models implemented in \cite{zhang2021pedestrian} for pedestrians while using a Kalman Filter state estimation. In this work, the vehicle's motion is modelled using a constant turn rate and constant velocity formulation. In their model, the two agents' motions are made coupled by transforming the pedestrians' coordinate into the vehicle's moving frame \cite{zhang2021pedestrian}.

\subsubsection{\textbf{Heuristic-based prediction methods}}
There are other expert-based motion prediction approaches where the hand-crafted decision rule for generating the next state from the current state of the pedestrian and its neighbours is based on some heuristics. These heuristics are often related to avoiding collisions while remaining close to the original track (e.g., \cite{anvari2015modelling,jan2020self}). Such heuristics are often used to form a utility function that is tried to get maximized when solving for the next state
\cite{anvari2015modelling,jan2020self,johora2018modeling,johora2021transferability,hossain2020conceptual}. In this line, Anvari et al. searched for the speed and heading change that besides resolving a pedestrian's conflict with a vehicle will result in the minimum deviation from the original path to the destination \cite{anvari2015modelling}. These hand-crafted rules are constructed based on some heuristics. For example, the heading angle (i.e., the direction in which the pedestrian's path is pointing) in \cite{jan2020self} is decided based on the straightest direction to the goal position while accounting for the distance to the closest collision in that direction. Then the speed of the pedestrian in the selected direction is calculated based on a specified minimum time to collision value. 

Another human-crafted rule-based motion prediction approach is the game theory methods that have been used along with SFM in a couple of papers \cite{johora2018modeling,johora2021transferability,hossain2020conceptual}. In these game-theoretic approaches, a payoff matrix for pedestrians' and cars' few discrete actions are hand-crafted based on some heuristics. Using this payoff matrix the optimal strategies of the two interacting pedestrian and vehicle agents are found by solving the Nash equilibrium which results in the agent's predicted next state.

The Cellular Automota method (CA) is another approach commonly used in simulation environments for generating agent's trajectories as consecutive occupied cells over time in a discretized world \cite{cheng2019study,li2015studies}. The transition between different cells in CA is governed by some hand-crafted rules and pre-specified probabilities. 
In \cite{cheng2019study}, the pedestrians' motion in a drop-off area at a railway station is modelled using this method. In this work, the direction and speed of the pedestrian are determined based on rules including the gap acceptance of pedestrians which is defined as the pedestrian's minimum comfortable time left for an approaching vehicle to reach the intersection point when a pedestrian starts crossing in front of the vehicle~\cite{cheng2019study}.

\subsection{Data-driven Models}
Instead of relying on explicitly defined dynamic equations or hand-crafted rules for describing the pedestrians' motions, many researchers have worked on learning these behavioural motion patterns from data (e.g., \cite{rasouli2019pie,sun2019interactive,anderson2020off}). This approach is followed in the literature with the goal of modelling pedestrians' more complex movements that cannot be captured by a hand-tuned function. For this purpose, different function approximates have been fitted to the human trajectory data ranging from linear regression models (e.g. \cite{kabtoul2020towards,girase2021loki}) to more complex deep learning models (e.g., \cite{bi2019joint,zhang2022learning,li2021interactive}). We identified four categories of data-driven models: Deep learning, Bayesian Networks, Linear model, and Reinforcement learning-based models are discussed in the following subsections.

\subsubsection{\textbf{Deep learning models}}In deep learning methods for pedestrian motion modelling, a highly non-linear function represented by a neural network is used for predicting the pedestrian's future state given their trajectory history. Due to the sequential nature of pedestrian trajectory, many of the models proposed in this area are based on recurrent neural networks (RNN) that can capture the dependencies between time series data. Different kinds of RNNs are used in the literature such as the Long Short Term Memory (LSTM) network~\cite{bi2019joint,cheng2018modeling} or the Gated Recurrent Unit (GRU)~\cite{girase2021loki}. The output of the popularly used LSTM cells, called the \textit{hidden state} encodes the trajectory history of each agent.

For the purpose of trajectory prediction, usually, an encoder-decoder architecture is used~\cite{rasouli2019pie,kim2020pedestrian,dos2021pedestrian,ridel2019understanding} where the states of the pedestrian, mainly consisting of their position coordinates will first get encoded into a hidden state as part of the encoder module. Then the extracted feature of each pedestrian's trajectory along with all other related features (e.g., the hidden states of other agents or scene features) will go through a decoder module to output the predicted states of the pedestrian's future trajectory. Other encoded features used for the trajectory prediction of a pedestrian in the literature are the pedestrian's pose~\cite{dos2021pedestrian,chen2020pedestrian} or head orientation~\cite{ridel2019understanding} as a representation of their intentions, other agents encoded trajectories or motion features for interaction modelling~\cite{zhang2022learning,bi2019joint} and also scene features for considering agent-environment context~\cite{cheng2020mcenet}. The way the encoded features of different agents are shared with each other in the architecture for modelling interaction is discussed in more detail in section \ref{Q2}. As another method, a transformer network is used in~\cite{yin2021multimodal} for pedestrian trajectory prediction.

While the above methods apply sequential models, it should be pointed out that non-sequential models such as the Convolutional Neural Networks (CNN) have also been used for trajectory prediction~\cite{zhang2022learning,cai2021pedestrian}. CNN is used in~\cite{cai2021pedestrian} for extracting trajectory patterns and used for predicting the next piece of trajectory among the alternative paths within a range of possible smooth turns.

Others have used CNN in combination with LSTM to capture the dependencies between the spatial features in form of agent-agent and agent-scene interaction along with the temporal features encoded by LSTM~\cite{chen2021modeling,wang2021multi,chandra2019robusttp,cheng2020mcenet,chen2020pedestrian,chandra2019traphic}. These CNN are therefore part of the module that encodes the interaction effects and will be discussed later in section \ref{Q2}.

An alternative approach followed in the literature for capturing spatial and temporal dependencies simultaneously in a trajectory prediction problem is the use of Graph Neural Networks (GNN)~\cite{li2021interactive,carrasco2021scout,mo2022multi,ma2019trafficpredict,zhang2021probabilistic,li2021hierarchical}. In the structure of a GNN, each agent is modelled with a node in the graph and there exist two types of edges connecting the nodes: temporal edges and spatial edges. Therefore, these models are also called Spatio-Temporal graphs \cite{li2021spatio,zhang2021probabilistic,wang2021multi}. The temporal edges model the relation between each agent’s individual node attributes over time. The spatial edges which connect the nodes of different agents together model the interaction between the agents and will be discussed further in section \ref{Q2}. The attributes of the edges and nodes are embedded in a feature vector and inputted to an LSTM to encode their information over time. These data are then concatenated for predicting each node's future trajectory.

A variant of GNNs called Graph Convolutional Networks (GCN) has also been used for trajectory prediction of pedestrians in a mixed environment of multi-class agents~\cite{men2022pytorch,rainbow2021semantics,su2022trajectory}. In these models, the feature of each node and the structure of the graph's connections in form of an adjacency matrix are inputted to a multi-layer convolutional network for extraction of common patterns in the trajectories.  

Creating multiple plausible future trajectories through repeated sampling of generative models is another common approach followed in the literature for considering the multi-modality nature of human trajectory. This is done through the use of Generative Adversarial Networks (GAN) in~\cite{zuo2021map,hassan2021predicting,eiffert2020probabilistic,lai2020trajectory,wang2021multi,hu2020collaborative} with different architecture for the generator and the discriminator but all consisting of LSTM layers. The other generative model used in the literature is the Conditional Variational Autoencoders (CVAE) which formulates the future trajectory of each agent conditioned on its past trajectory and a latent variable that can be sampled multiple times~\cite{wang2021ltn,herman2021pedestrian,girase2021loki,bhattacharyya2021euro,cheng2020mcenet,zhang2021probabilistic}. This latent variable is learned during the training phase by providing both the trajectory history and the future trajectory of an agent.

\subsubsection{\textbf{Dynamic Bayesian Networks}}
A Dynamic Bayesian Network (DBN) has been used in~\cite{sun2019interactive} for modelling the state change of multiple heterogeneous agents. Two main tasks of intention estimation and trajectory prediction are performed simultaneously in their proposed model. The clear causal dependencies between the variables in the Bayesian graphical model can bring more interpretability to the DBN approaches compared to the deep learning methods. The variables of the proposed DBN in \cite{sun2019interactive} include agents' continuous state, action, observable state and a discrete latent state used for analyzing the decision and actions of agents. The approximate inference for estimating the intentions and predicting the trajectories of agents is performed through the use of a particle filter and the PedX dataset\cite{kim2019pedx}.

\subsubsection{\textbf{Linear regression models}}
A trajectory planning model is proposed in \cite{kabtoul2020towards} that predicts the velocity of the pedestrian at the next time step as a linear function of its current velocity and a vector of motion parameters consisting of the vehicle's influence on the pedestrian, the pedestrian's own goal position, the effect of surrounding pedestrians, and the cooperative factor of the pedestrian. Assuming that the pedestrian's velocity is linearly influenced by all these factors, the parameters of the function that maps these inputs to the predicted output is found through linear regression and using the CITR dataset \cite{yang2019top}. The defined cooperation factor in the formulation itself is also modelled as a linear function of parameters such as the probability of collision, deformation of the personal zone and the density of space surrounding the agent, and the parameters of this model are also derived using linear regression.  

In another method proposed in \cite{anderson2020off}, a pedestrian-vehicle interaction model is used on top of a random walk velocity model (i.e., the velocity of the next time step being normally distributed with a mean equal to the velocity of the current time step). In this model, first, the pedestrian specifies the vehicles that require attention and whether the pedestrian should yield to the vehicle or continue at their own desired speed. These decisions are made based on a defined piece-wise linear risk function knowing the closest distance and the time to the closest distance between the pedestrian and the vehicle. Once the yielding decision is made, the pedestrian velocity is predicted based on another vehicle influence function that is also piece-wise linear and determines the fraction of the desired speed during the yielding \cite{anderson2020off}. Therefore, the pedestrians' yielding decision as well as their next step velocity, is modelled to be a linear function of parameters such as lateral distance between pedestrian and vehicle. The coefficients of these functions are then learned using the DUT datasets \cite{yang2019top}.

\subsubsection{\textbf{Reinforcement learning Methods}}
The collision-free trajectory of multiple heterogeneous agents can be generated simultaneously using multi-agent planning methods such as the multi-agent deep reinforcement learning (RL) method implemented in \cite{nasernejad2021modeling,nasernejad2022multiagent}. The collision avoidance between a pedestrian and a vehicle is modelled as a Markov Game in \cite{nasernejad2022multiagent} and the optimal policy of the agent is found using an actor-critic network in the RL framework. Pedestrians' and vehicles' reward functions are recovered using a multi-agent adversarial inverse RL approach~\cite{nasernejad2021modeling,nasernejad2022multiagent}. 

\subsection{Combination of Data-driven and Expert-based Models:}

In a few paper, deep learning models have been combined with expert-based models for predicting pedestrians' future trajectories \cite{johora2020agent,li2021rain,li2021spatio}. Johora et al. \cite{johora2020agent}, have shown that the combined approach outperforms each of the pure methods. In their architecture, the expert-based methods consist of the three interacting modules of free-flow trajectory planning, the social force model, and the game-theory decision layer. This combination is executed in parallel with the deep learning module. The output of the deep learning-based module is then used when no conflict between the predicted trajectories of agents is detected. In case of a detected conflict in the generative trajectories of the deep learning model, the output of the Expert-based model will be implemented.

Moreover, in the data-driven deep learning models proposed in \cite{li2021rain,li2021spatio}, some physics-based dynamic constraints on the motion of agents are applied at the near-end layers of the model for guaranteeing the feasibility of the generated trajectories. 
In \cite{li2021spatio}, the actions of agents are predicted through the network and a kinematic cell that takes as input that action and the current state, generates as output the next state of the agent.  

In a couple of expert-based models (e.g., SFM), the parameters of the model are estimated by using real-world trajectory data \cite{anvari2015modelling,johora2021transferability,predhumeau2022agent,yang2020social,cheng2019study}. In these methods, parameters such as safety distances, range of repulsive forces, payoff matrices of the game \cite{johora2021transferability} or the social factor and perception zones in the SFM \cite{predhumeau2022agent} are calibrated through a process to get the minimum difference between the ground-truth trajectories in the dataset and the simulated ones using the model. 

\subsection{\textbf{Discussion on Strength and Weaknesses of Expert-based and Data-driven Models}}

Expert-based approaches, as described, require human hand-crafted explicit decision rules of how pedestrians' trajectories evolve over time. These designed transition rules as usually not environment-specific, can be used across multiple environments without any need for training data which makes them suitable for unstructured environments. However, these models might not be able to capture all the complexity of the real world pedestrian motion. In addition, the intervention of human modellers in drafting general reasoning principles makes it hard to scale up these models for new larger problems while taking into account all the different factors influencing human's complex decision-making process.

On the other hand, data-driven approaches can capture the complex non-linear trajectory behaviour of pedestrians embedded in the training dataset with a great accuracy. Therefore, these methods are suitable for situations with complex unknown dynamical effect that is hard to be captured through hand-crafted formulations. However, the problem with deep learning methods is their black-box nature that makes the generated behaviours hard to explain. The lack of control over the generated trajectories could sometimes cause unjustifiable outputs. As these models are trained using real-world date they could get biased by some contextual information of the environment from which the trajectories are captured. Therefore, data-driven approaches require a sufficient amount of data from different sites and scenarios for satisfying the generalizability of the trained model. This is an important factor to consider when it comes to models trained for unstructured environments as diverse interaction scenarios can happen in these environments. Additionally, the computational cost of training deep-learning models could become a bottleneck in these approaches.

Cheng et al. \cite{cheng2021trajectory} have compared a proposed expert-based trajectory model with another designed deep-learning model for shared spaces. By using a common framework for a fair comparison, they showed that the expert-based model predicts collision-free trajectories with the downside of their predictions tending to be homogeneous. On the other hand, they showed that the deep learning approach, while being accurate for short-term predictions, may generate near-collision trajectories in the long term as the prediction accuracy degrades for long prediction horizons \cite{cheng2021trajectory}.

The above work is a comparison study between two specific example models from the two broad category of expert-based and data-drive-based approaches for trajectory prediction. However, more studies are required for comparing the prediction performance of different modelling approaches in general.

\section{RQ2: Interaction Modeling} \label{Q2}

The presence of a moving vehicle close to a pedestrian can highly affect the pedestrian’s motion behaviour and is an important factor to consider when modelling or predicting the pedestrian’s future trajectory~\cite{zhang2022learning,eiffert2020probabilistic}. This effect, referred to as the pedestrian-vehicle interaction effect~\cite{zhang2022learning}, has been modelled in many different ways in the literature depending on the type of model used for generating the pedestrians’ trajectory such as being expert-based or data-driven.

From a broad perspective, these interaction effects can be categorized to be modelled either explicitly or implicitly. In the explicit interaction modelling approaches (e.g., see~\cite{yang2017agent,anvari2015modelling,johora2018modeling,hesham2021advanced}), the effect of the vehicle on a pedestrian’s motion is forced through some clear terms in the formulation of the pedestrian’s motion such as through explicit forces in the social force model (e.g.,~\cite{predhumeau2022agent,zhang2022modeling}). On the other hand, data-driven modules often account for these interactions implicitly by using the vehicle’s trajectory as another input to the model along with the target pedestrian’s own trajectory (e.g., see~\cite{bi2019joint,carrasco2021scout}).
As these models are trained on datasets collected from real-world interaction scenarios, it is expected that the model will learn the interaction already embedded in the data.

\begin{figure}[b!]
  \centering
  \includegraphics[width=0.9\linewidth]{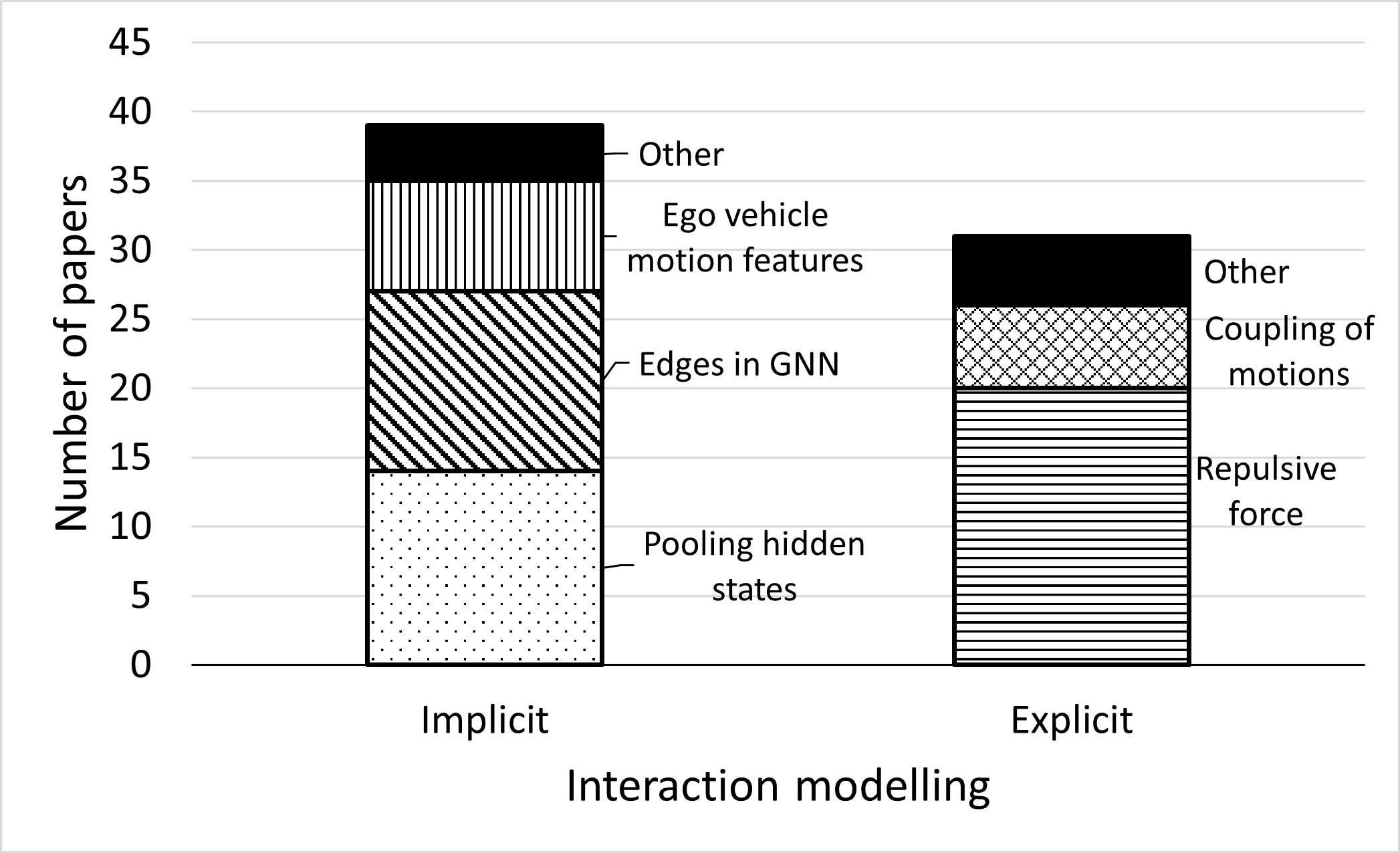}
  \caption{The different methods used for modelling pedestrian-vehicle interaction in the pedestrian's predicted motion. The number of reviewed papers in each category is shown.}
  \label{fig:Inter}
\end{figure}

More details on the interaction models proposed in the literature in each sub-category will be discussed in the following sections. Fig. \ref{fig:Inter} summarizes the number of reviewed papers in each category.

\subsection{Explicit Interaction Modeling}

We categorized explicit interaction approaches used in the literature into applying a) repulsive forces, b) combination of SFM with other collision avoidance strategies c) direct coupling of motions and d) other methods.

\subsubsection{\textbf{Repulsive forces}}
The original social force model (SFM), proposed by Helbing and Molnar (1995) \cite{helbing1995social}, only considers the social interactions between pedestrians. Many papers have extended that model by introducing additional forces for modelling pedestrian-vehicle interaction \cite{yang2018crowd,borsche2019microscopic}. The way these additional forces are formulated is the main difference between these extended social force models. 

In these models, each vehicle exerts a repulsive force on the pedestrian based on its relative distance and approach direction. The effect of the relative interaction distance is captured in the so-called \textit{decaying function} \cite{yang2020social} which is usually selected to be an exponential function of the distance \cite{yang2017agent,anvari2014long}. Another function which is used in some of the SFM model formulations is called the \textit{anisotropy function} \cite{yang2020social,yang2018social}. This function formulates the effect of the different interacting directions on the magnitude of the repulsive force. Such a function will consider, for example, the higher influence that a pedestrian walking toward the vehicle will experience compared to another pedestrian that is walking away from the vehicle when both of them are positioned at the same distance from the vehicle, which means having the same value for the decaying function \cite{yang2020social,yang2018social}.

Some papers have modelled the vehicle with a circular shape same as how pedestrians are modelled in SFM but with a larger radius~\cite{predhumeau2022agent,predhumeau2021agent}. In these models, the asymmetrical effect of the vehicle's influence is accounted for in the anisotropy function or considering the approach angle and its change rate~\cite{predhumeau2022agent,predhumeau2021agent}. On the other hand, some others have modelled a vehicle with an ellipse and considered the asymmetry of the vehicle’s influence by calculating the pedestrian’s relative distance based on its angular position with respect to the ellipse \cite{anvari2014long,zhang2022modeling,anvari2015modelling}.

In \cite{yang2017agent}, one foci of the ellipse is placed on the middle rear point of the vehicle and the other is placed on an extended front point depending on the vehicle's speed. In this way, the model is accounting for the more dangerous front zone of the vehicle compared to its rear area and also extending the danger zone as the vehicle's speed increases. A repulsive force perpendicular to the tangent line of the ellipse edge is experienced by a nearby pedestrian in their model \cite{yang2017agent}.

Others have used a fixed ellipse which encloses the vehicles and repulsive force is modelled as an exponential function of the distance between the vehicle and the pedestrian, where the distance depends on the relative orientation of the pedestrian with respect to the vehicle \cite{zhang2022modeling,anvari2014long,anvari2015modelling}. Yang et al, (2018) define three areas around the vehicle --- namely the front area, body area, and back area --- and propose a different formulation for the lateral and longitudinal forces of the vehicle on the pedestrian in each area \cite{yang2018social}. Of course, the front area always has the highest magnitude of forces due to being more dangerous. A rectangular shape contour with an extended triangular shape in the front is used for the vehicle in \cite{yang2020social}. The vehicle-pedestrian interaction force is calculated as a function of the minimum distance between the pedestrian to the contour. The magnitude of the force is also adjusted according to the walking direction of the pedestrian.

The interaction between a vehicle and pedestrians in very close contact is also modelled with these repulsive forces in \cite{janapalli2019heterogeneous,hesham2021advanced}. In a simulation case of an ambulance trying to pass through a crowded scene of pedestrians, a collision penalty force is exerted whenever the vehicle and the pedestrians penetrate into each other's personal space. This repulsive force will help the agent regain their personal space when it is violated by others. The personal space for the vehicle is represented with 3D cones around the polygon shape of the vehicle and pedestrians are modelled with circles.

\subsubsection{\textbf{Combination of SFM with other collision avoidance strategies}}

The repulsive force in the SFM increases when the agents get closer to each other and almost vanishes when they get far away \cite{helbing1995social}. Because of this, short-range repulsive forces of the SFM can cause frequent urgent detours when it comes to pedestrian-vehicle collision avoidance. Therefore, many papers have combined the SFM with other collision avoidance strategies~\cite{anvari2014long,anvari2015modelling,rinke2017multi,predhumeau2022agent,predhumeau2021agent}. In this regard, a long-range collision avoidance method is proposed in \cite{anvari2014long,anvari2015modelling} for handling potential collisions that might last by just following the SFM formulation. Within this method, potential conflicts are predicted by geometrically projecting the shadow of the pedestrian in the direction of the car. The minimum speed and direction change for avoiding the collision are then calculated by solving an optimization problem.  

For long-range conflicts, defined as conflicts happening in more than 2 seconds in \cite{rinke2017multi}, the authors consider a force that keeps the pedestrian in a safe zone. This is done by defining a force perpendicular to the car which models the pedestrian’s tendency to walk parallel to the car in a shared space environment. 

SFM is used alongside a proposed decision model in \cite{predhumeau2022agent,predhumeau2021agent}, which gets activated whenever a conflict is predicted with the vehicle by referring to the time-to-collision (TTC) parameter. The decision model decides on an action based on (a) the type of the interaction being from the front, back, or lateral, and (b) the expected crossing order between the pedestrian and the vehicle. These actions are implemented through some newly introduced forces in the SFM and can create a sharp turn, a stop, or a step-back action. For no-conflicting interactions, the simple SFM in which the vehicle just has a repulsive force on the pedestrian is used.

SFM's ability to directly map perception to action is used in \cite{johora2018modeling,johora2020zone,johora2021transferability} for solving simple reactive interactions. However, for handling more complicated interactions which require deciding among different alternative actions, a game-theoretic layer is added on top of the SFM. 
The decided action at the game-theoretic layer is executed by the force-based layer. This model specifically designed for shared space was named Game-Theoretic Social Force Model (GSFM) \cite{johora2021transferability}. For handling multiple conflicts in a shared space more efficiently using this model, the authors proposed the concept of using a conflict graph \cite{hossain2020conceptual}. This graph will keep track of the conflicts at each time step and helps the model prioritize the conflicts based on their urgency instead of solving each detected conflict right away. Later, the GSFM is proposed to be used beside a deep learning model and perform as a safe backup for the data-driven model \cite{johora2020agent}. Within this structure, whenever the predicted trajectories of the data-driven model end up in a collision, the output of the GSFM will be used instead as a guaranteed collision-free trajectory prediction~\cite{johora2020agent}.

\subsubsection{\textbf{Direct coupling of motions}}

Interactions could also be modelled explicitly by coupling the vehicle and the pedestrian’s motion equations and directly considering the effect of an agent's action on the other's motion decisions. In this regard, Zhang et al. (2021) \cite{zhang2021pedestrian}, have used a constant turn rate and velocity model (CTRV) for modelling the vehicle’s motion. Using the corresponding equations, they mapped the pedestrian's state to ego vehicle's coordinate. Using a constant velocity model for the pedestrian’s motion in the vehicle's frame, the authors derived a coupled system of equations in which the pedestrian’s position and velocity are the states of the systems and the ego-vehicle velocity are used as control variables~\cite{zhang2021pedestrian}. 

Game theoretic methods used in the literature for considering the effect of interacting agents' actions on one another also fall into this category as they solve for a coupled decision for the interacting agents by searching the Nash equilibrium \cite{johora2018modeling,johora2020zone,johora2021transferability,hossain2020conceptual,johora2020agent}. These calculations are based on a payoff matrix that explicitly encodes the effect of one agent's decision on the other. For example, the game-theoretic layer in the proposed architectures of \cite{johora2018modeling,johora2020zone,johora2021transferability}, models the interaction between a pedestrian and a vehicle as a Stackelberg sequential leader-follower game \cite{von2010market} where a payoff matrix is designed for possible discrete actions of \textit{continue}, \textit{decelerate}, or \textit{deviate} for the pedestrian, and \textit{continue} and \textit{decelerate} for the vehicle. The final optimal actions are executed via the social force layer to generate the coupled trajectories.

\subsubsection{\textbf{Other explicit interaction models}}

There are also other methods used to explicitly account for the effect of the vehicle on the pedestrians' future states. In \cite{jan2020self} at each time step, the speed and direction of the pedestrian are chosen in a way to guarantee no collision with the vehicle when the trajectories cross. The effect of the vehicle on the pedestrian's speed is also modelled in \cite{anderson2020off} based on the calculation of the collision risk which is defined to be a function of the minimum distance and the time to the minimum distance in each interaction.

In the cellular Automata formulation in \cite{cheng2019study,li2015studies} pedestrians decide to cross in front of the vehicle if the remaining time left for the vehicle to get to the position of the pedestrian is higher than a threshold. This common criterion has been used in the literature, especially for modelling pedestrian crossing decisions on streets and is called gap acceptance of the pedestrian. But Cheng et al. (2010)~\cite{cheng2019study} used the same concept for the drop-off area of a railway station which was more similar to an unstructured environment. Also, in \cite{li2015studies} when a vehicle suddenly runs into the critical gap of a pedestrian, the person's reaction is simulated among four possible behaviour with a probability assigned to each: Going forward, rushing, waiting, going backward.

Other heuristics such as TTC have also been used in combination with the Social force model \cite{zhang2022modeling,predhumeau2021agent} to keep track of interaction between pedestrians and vehicles. In \cite{zhang2022modeling} the TTC governs the principal behind the vehicles behaviour in interaction with pedestrians, where the vehicle breaks whenever the TTC is lower than a threshold. In \cite{predhumeau2021agent} another decision layer on top of the SFM gets activated when the TTC between the agents gets below a threshold.

A cooperation factor for pedestrians is defined in \cite{kabtoul2020towards} as an explicit interaction element between a pedestrian and a vehicle close to each other. This cooperation factor is a function of the probability of collision between the two agents and the deformation of the pedestrian's personal zone. Both of these features account for the effect of the vehicle's motion on the pedestrian's future state. In the trajectory planning part of this model, the next step velocity of the pedestrian is predicted as a function of the current velocity and a 7-dimensional parameter. These motion parameters encode the effect of the vehicle and the pedestrian's destination weighted through the cooperation factor, along with two other terms related to the effect of surrounding pedestrians and the change in the agent's cooperation. The parameters of the linear regression models in this paper are estimated using real data \cite{kabtoul2020towards}.

\subsection{Implicit Interaction Modeling}

In data-driven deep learning methods for pedestrian trajectory prediction, interaction effects are usually encoded through a separate module that its output is used along with the agent's own encoded trajectory history to generate the predicted next state of the agent. The module that encodes the interaction effects uses the trajectory information of nearby agents along with the target agent's own information for implicitly learning the effect of interacting neighbours on the target-agent trajectory from real data.

Different models have been proposed for combining the trajectories of different agents in the interaction module such as pooling mechanisms or graph neural networks. Some papers that are focused on predicting the trajectory of a single pedestrian from the egocentric view of a moving vehicle, try to account for the interaction between the pedestrian and the ego vehicle using some moving features from the vehicle in the data-driven prediction model \cite{quan2021holistic,rasouli2019pie,dos2021pedestrian,yin2021multimodal,rasouli2021bifold}. The interaction formulation in each of these three models is discussed separately in the following subsections.  

\subsubsection{\textbf{Pooling trajectories of interacting agents}}

One of the common ways for pooling hidden states of all surrounding agents is by constructing an occupancy grid map centred at the target agent's position \cite{bi2019joint, hassan2021predicting, cheng2020mcenet}. In these occupancy maps, the hidden state of all agents residing in the same grid cell will be summed together, building a tensor that embeds information of all interacting agents that can impact the future trajectory of the target pedestrian. This tensor input will be used along with the target agent's own spatial hidden state as the two main inputs to the LSTM network used for the trajectory prediction.

While the above-mentioned grid map used for constructing the tensor input has a rectangular shape in\cite{bi2019joint, hassan2021predicting}, Cheng et al, have used a circular polar occupancy map where the orientation and the distance of the agents with respect to the target pedestrian specify the occupied cells \cite{cheng2020mcenet}. Moreover, this paper also pays attention to pedestrians that walk in groups and the interaction differences between group and non-group members. Therefore, the agent's group members are not included in the construction of the person's occupancy map. 

In the above-mentioned papers, grid maps used for social pooling are constructed locally around each target agent. A global version of such spatial feature maps at each time step is proposed and used in \cite{wang2021multi} by dividing the bird's-eye view of the scene into discrete cells. In this constructed map the feature encoding of each agent is placed in a tensor based on the agent's position coordinate. These 2D tensors at each time step go through a CNN while the temporal dependencies between these spatial maps over time are also analyzed by using a separate LSTM network.    

Two maps are created for each agent in \cite{chandra2019traphic} called the \textit{Horizon map}, and the \textit{Neighbour map} by pooling the embedding of nearby agents in two defined regions. The horizon region, which contains the agent's prioritized interactions with others has a semi-elliptical form in front of the agent while the neighbourhood region is defined as a larger elliptical area around the target agent with a fixed number of closest nearby agents in that region being considered as neighbours.
Each of these created maps goes through a Convolutional network that can capture the local dependencies between features of interacting agents in that map and the outputs are then concatenated with the target agents embedding to predict the future trajectory of the target agent \cite{chandra2019traphic, chandra2019robusttp}.

While the construction of a tensor input based on an occupancy grid map for encoding interactions are mainly based on relative distances \cite{bi2019joint, hassan2021predicting, cheng2020mcenet}, Cheng et al, have proposed a tensor input that relies on collision probability with other agents \cite{cheng2018modeling}. However, their collision probability is again formulated as a function of the distance between the pedestrian and the vehicle considering their approach direction and the elliptical shape of the car. 

Instead of using occupancy grid maps and storing the hidden states of nearby agents in such a map in form of a tensor for keeping the relative positions, others have used max-pooling or softmax pooling layers to sum the hidden state of all surrounding agents. In these softmax layers the weighted sum of the neighbouring agent's hidden state is calculated with a weighting that is a function of the distance between the target agent and all the other nearby agents \cite{eiffert2020probabilistic}. In other words, these papers directly use the value of the relative distance instead of accounting for that based on the position of the pedestrians on an occupancy map.

In this line, The pedestrian-vehicle interaction weights are calculated based on their relative distances in \cite{zhang2022learning} using a max pooling layer. These weights are then aggregated with the vehicles' embedded moving state to get the pedestrian-vehicle interaction feature.

In \cite{eiffert2020probabilistic} the output of the softmax layer is used for aggregating the interaction effect of multiple neighbouring pedestrians by learning an importance weight for each. The input to their softmax layer for each surrounding pedestrian is an embedding of both the target pedestrian's distance from that neighbouring agent and from the vehicle. Therefore, the vehicle's effect is seen through its impact on pedestrian-pedestrian interactions \cite{eiffert2020probabilistic}.

Two different types of pooling modules are proposed in \cite{zuo2021map}, one for homogeneous agents and the other for heterogeneous agents. The pedestrian-vehicle interaction effect is considered in the heterogeneous pooling module in which the average output of the vehicle's homogeneous pooling module is combined with each pedestrian's output from its own homogeneous pooling module. 

The relative importance of surrounding agents when aggregating their state can also be specified with an attention mechanism \cite{wang2021ltn}. While the attention coefficients of the surrounding agents are mainly constructed based on their relative Euclidean distance to the target pedestrian, other social features such as the bearing angle between the two agents and their speculation distance were also used \cite{lai2020trajectory}. In fact, in \cite{lai2020trajectory} two different attention coefficient is defined. One considers all the agents in the scene even if they are far away from the target agent and the other takes into account only closer agents that are located inside a semicircular region in front of the target agent. The latter helps the network to put more emphasis on interaction in front of the target agent. These two weightings have been added together to construct the final attention scales for each agent.

Unlike the above methods that use position coordinates as input and pool together the hidden states of agents' position over time, a couple of models in this area have used the image inputs and conducted pooling on the features extracted from images \cite{rasouli2021bifold,chen2021modeling,chen2020pedestrian}. In \cite{rasouli2021bifold} the images captured from an ego-centric view are used as input. In their method, each image is divided into different categories including the target pedestrian, surrounding pedestrians, vehicles and static context and each extracted categorical image is processed separately with convolutional layers (CNN) and an LSTM. The outputs of all these categorical images are then pooled together using an attention mechanism to account for the interaction between these agents.

The methods proposed in \cite{chen2021modeling,chen2020pedestrian}, work on the images captured from a stationary camera. A grid map around each pedestrian in the image is constructed and the relative position of 13 object categories including vehicles located in the pedestrian's grid map is encoded in a multi-channel tensor (one channel for each category). This tensor input is then combined with other inputs and goes through multiple convolutional and LSTM layers to encode interaction between the agent in different categories for predicting the next position of the agent in the image.

\subsubsection{\textbf{Message passing through spatial} edges of graph neural network}

In the structure of the graph neural networks (GNN) used for pedestrian trajectory prediction, the spatial edges which connect together different nodes, model the interaction between the agents representing those nodes and the effect of these interactions on the future position of each agent.
The process of aggregating information from connected nodes in a graph and using that for updating the node's representation is referred to as message passing. This mechanism is used for formulating the effect of interacting agents on a target pedestrian's motion in GNNs. An attention mechanism is usually used in these models as a way of encoding the relative importance of the connected edges for the target agent.

Two nodes or agents are connected through an edge in the graph and are considered to be interacting with each other based on a criterion that varies between different papers. A popular criterion is the spatial distance where two agents are connected with a spatial edge in the graph if they are within a certain distance from each other \cite{carrasco2021scout, li2021spatio, girase2021loki}. Then the features of connected neighbours to an agent will be used with the agent’s own feature for the calculation of its future location. While there are papers that rely on a criterion for specifying connecting edges, some just start with a fully connected graph \cite{hu2020collaborative}, in other words accounting for all agents in the scene. 

 In \cite{li2021rain} the existence of these edges is decided based on a reinforcement learning framework where the actions are specified as making each edge on or off and the reward is defined according to how good the overall trajectory prediction is for the selected graph connections. They refer to this process as the hard attention module of their overall framework while also having soft graph attention which weights the relative importance of the existing edges to prioritize interactions that have the highest impact on the future position of each agent. 

 There are a couple of papers that used directed graph \cite{mo2022multi, su2022trajectory, girase2021loki, hu2020collaborative, zhang2021probabilistic} as opposed to undirected ones. Directed graphs are usually used to account for the asymmetry of the interactions \cite{su2022trajectory}. The interactions can be considered asymmetric by accounting for the effects such as field of view, moving direction and moving speed of agents \cite{su2022trajectory}.  In this regard, three different directed graph with different logic for node connections is proposed in \cite{su2022trajectory}, namely: \textit{view graph}, \textit{direction graph} and \textit{rate graph}. In the \textit{view graph} agents are connected to neighbours that are within their field of view which depends on each agent's moving direction. For the \textit{direction graph} the agents are assumed to pay attention to those that might have a potential conflict with the agent by considering the agents moving directions and finding the possibility of crossing paths. Finally, in the \textit{rate graph} the effect of neighbours’ speed on the amount that they will gain attention is considered for constructing the edges of the graph.

In \cite{girase2021loki}, the graph and its encoded interactions are also used for predicting the short-term intentions of agents in the form of a probability distribution function over a discrete set of intentions~\cite{girase2021loki}. This predicted intention is then used along with the graph structure itself for predicting each agent’s future trajectory.

Some have used Graph Convolutional Network (GCN) to directly operate on the graph \cite{carrasco2021scout,men2022pytorch,rainbow2021semantics, su2022trajectory}. In these models, an adjacency matrix is constructed which stores the connected edges in the graph and specifies weights for them that are proportional to the inverse of the relative speed \cite{rainbow2021semantics} or relative distances \cite{carrasco2021scout,su2022trajectory} of the interacting agents. The justification behind this selected function is that further away neighbours are less likely to have an impact on an agent’s future trajectory. In the structure of a GCN, the attributes of the connected nodes will be aggregated using the adjacency matrix and combined with the agent’s own node attribute to be used for predicting the agent’s future path. Therefore, the node’s connectivity or in other words the agent’s interactions with others will be considered in the prediction process.

As opposed to GCN, other GNN methods have used recurrent neural networks (RNN) such as LSTMs to encode the sequential dynamic features of the edges in the graph.  In the later, the attributes of the edge are relative measurements such as relative distance between the two connected traffic agents in the graph \cite{li2021interactive,ma2019trafficpredict,zhang2021probabilistic,mo2022multi,li2021spatio,girase2021loki,li2021hierarchical, hu2020collaborative}, relative velocity \cite{mo2022multi,li2021spatio,girase2021loki} or relative heading/yaw angel \cite{mo2022multi,li2021spatio}. These edge features are sent to an RNN (e.g., LSTM \cite{li2021interactive,zhang2021probabilistic,cai2021pedestrian}, GRU \cite{mo2022multi,li2021spatio}) to encode the information of the spatial interactions between the agents and to be used along other embedded information (e.g., the temporal edge) to output the predicted trajectory in subsequent layers.

Other features used as edge attributes in literature are the type of interacting agents and the label of the interactive event. The type of agents interacting with each other is represented by a unique encoder in \cite{li2021interactive, ma2019trafficpredict,li2021hierarchical} and used as a feature of the spatial edge.  Mo et al 
\cite{mo2022multi} also used edge type (a discrete indicator) as an edge attribute which is specified based on the type of agents interacting with each other. 

Another feature used for the edges of the GNN is interactive events \cite{li2021interactive,li2021hierarchical} that for a road interaction can be events such as overtaking from left, driving away from left or parallel driving \cite{li2021interactive}. In \cite{li2021hierarchical} a separate module is developed for interactive event recognition and is used as another way of modelling interactive behaviours besides the trajectory interactions encoded in the graph neural network. 

\subsubsection{\textbf{Ego vehicle-pedestrian interaction}}

For the papers that study the pedestrian’s trajectory prediction from an egocentric view, the interaction of the pedestrian with the ego-vehicle is the pedestrian-vehicle interaction considered in these papers. These interactions are usually modelled by inputting some characteristics of the ego vehicle's motion alongside the pedestrian’s own position sequences. One of the usual features used for this purpose is the ego vehicle's speed which could have a high effect on the decision and motion of the pedestrian that is interacting with the vehicle.

The ego vehicle's speed is used in a couple of papers for predicting the pedestrian's next position in the camera’s image \cite{quan2021holistic,rasouli2019pie,dos2021pedestrian,yin2021multimodal,rasouli2021bifold}. Some have even proposed the use of a separate network for predicting the ego vehicle's next speed and using that for trajectory prediction of the pedestrians \cite{rasouli2019pie,dos2021pedestrian}. Other papers have used features like the relative distance of the pedestrian to the ego-vehicle \cite{herman2021pedestrian} or the ego vehicle's location coordinates \cite{ridel2019understanding} along with the other pedestrian’s motion features. The pedestrian head orientation \cite{herman2021pedestrian} and their body pose are other interaction features used as input which could embed the pedestrians’ awareness of the vehicle or their intention to cross in front of the vehicle. 

While all the above papers are focused on pedestrian’s trajectory prediction and its interaction modelling from the perspective of an ego vehicle, Kim et al \cite{kim2020pedestrian}, have added the pedestrian’s perspective to this and considered interaction features like relative positions between the pedestrian and the vehicle, the pedestrian’s head orientation with respect to the vehicle and the vehicle’s speed. These features are used as interaction indicators when predicting the motion decisions from the pedestrian’s perspective and in the next stage are combined with the pedestrian’s predicted trajectory from the driver’s perspective, with its own considered interaction features, to generate the final motion forecast.    

However, capturing the scene from the perspective of an ego-vehicle means that all the pedestrians' motion sequences used in the above papers are with respect to a moving frame or in other words relative positions. Therefore, the use of vehicle features such as location or speed as another input could only be considered as a way of correcting for the effects of a moving frame and not necessarily an interaction consideration in the model. This is made clearer in \cite{cai2021pedestrian} where they used a relative motion transformation module to transfer the pedestrian's relative coordinate in the camera’s view to the absolute motion on the ground using the ego-motion of the vehicle-mounted camera. But even in this paper, the authors claim that with this module the effect of the vehicle’s ego-motion on pedestrians’ speed and direction changes for avoiding any collision with the vehicle will be encoded in their model.

\subsubsection{\textbf{Other implicit interaction modelling}}

In very few papers the trajectory prediction problem is modelled as a Bayesian network and the interactions are formulated through the causal and temporal connections in the graphical model \cite{sun2019interactive}. In the graphical model proposed in \cite{sun2019interactive}, the next state of each agent is predicted as a function of its previous state and action. The variables that decide the actions are the current state and the latent state Z. The latent state of each agent is conditioned on the state of all nearby agents. This is how the model accounts for interaction. 

The concept of using such latent variables that accounts for the interaction by taking as input all the agent' trajectories or states is also used \cite{bhattacharyya2021euro}. But instead of using a Bayesian network, they model the conditional probabilities as neural networks by using the structure of a variational auto-encoder (VAE). In the model proposed in \cite{bhattacharyya2021euro}, an attention mechanism is used to find the dependence of this latent variable on the agent's previous trajectories.

In \cite{nasernejad2022multiagent, nasernejad2021modeling}, the trajectory of a pedestrian and a vehicle while interacting with each other is generated using a reinforcement learning framework where each agent decides on its action based on the state of itself and the state of the other agent. In this paper, the reward function of the agents for collision avoidance is recovered using an inverse reinforcement learning (IRL) approach. Among the inputs that IRL algorithm uses are longitudinal and lateral distance as well as heading angle and speed differences between the vehicle and the pedestrian. Therefore, by being trained on real data a reward function is learned that implicitly considers the interaction between the pedestrian and the vehicle \cite{nasernejad2022multiagent, nasernejad2021modeling}.

\section{RQ3: Difference between pedestrian-pedestrian and pedestrian-vehicle Interaction} \label{Q3}

When modelling pedestrians' motion considering influences from both vehicles and other pedestrians, it is important to understand how pedestrian-vehicle interaction is implemented and how it differs from a pedestrian-pedestrian interaction. These differences are discussed here while being categorized based on the applied interaction model.

\subsection{Interaction of Agents in SFM}

In the Social Force Model where interactions are modelled through forces, the repulsive force exerted on the target pedestrian from another pedestrian is formulated differently from the repulsive force from a vehicle \cite{yang2017agent,yang2018crowd}. In some models, this is done through the selection of different values for some of the parameters in the formulation, such as the \textit{interactive strength} and \textit{interaction range} of the repulsive force from a vehicle compared to a pedestrian~\cite{borsche2019microscopic,johora2018modeling,johora2021transferability,anvari2014long}. Other parameters such as the social force factor, the perception zone and the preferred speed are given different values for a pedestrian versus a vehicle in \cite{predhumeau2021agent}. Moreover, the different field of view of vehicles compared to pedestrians are accounted for in \cite{johora2020zone} considering that only interactions with surrounding pedestrians and vehicles that fall into an agent's field of view will affect that agent's path. Some even solve pedestrian-vehicle conflicts differently compared to pedestrian-pedestrian conflicts by adding extra forces for pedestrian-vehicle interactions. These extra forces are decided based on other layers added in the model, on top of the SFM, such as game theory \cite{johora2020zone,hossain2020conceptual} or other designed decision models \cite{predhumeau2021agent,predhumeau2022agent}.

The other difference in the interaction formulation of a Social Force Model is caused by considering the difference in geometry and size of the vehicle compared to a pedestrian, which affects the calculation of the distance between the two agents \cite{anvari2014long,zhang2022modeling,anvari2015modelling}. The size of the agent also impacts the personal space (e.g., the space around each agent into which any encroachment feels uncomfortable) \cite{hesham2021advanced,janapalli2019heterogeneous} and conflict zone (e.g., A distance below which there is a risk of collision)\cite{predhumeau2021agent,predhumeau2022agent} of a pedestrian versus a vehicle. While pedestrians are basically considered to have a circular shape in SFM, vehicles are commonly modelled as an ellipse. Due to this elliptical shape, the distance between a pedestrian and a vehicle also depends on the orientation of the vehicle compared to the pedestrian \cite{anvari2014long,zhang2022modeling,anvari2015modelling}. The orientation of the vehicle and the shape of the danger zone around it which is made a function of the vehicle's speed in \cite{yang2017agent,yang2020social} will affect the magnitude and the direction of the repulsive force exerted by the vehicle on the pedestrian \cite{yang2017agent,yang2020social,yang2018social}.

\subsection{Interaction of Agents in Models that use a Pooling Method}

In articles that model the interactions implicitly by pooling the hidden states of different agents, a distinction between pedestrian-pedestrian versus pedestrian-vehicle interaction is made by having different pooling modules \cite{zhang2022learning,zuo2021map,chen2021modeling,bi2019joint,chen2020pedestrian}. This is done in~\cite{bi2019joint} by having two separate occupancy maps for each pedestrian, one for pooling the hidden states of nearby pedestrians and the other for nearby vehicles. The output of both pooling modules is concatenated with the pedestrian's own hidden state for predicting its next state. The same concept is followed in \cite{chen2021modeling,chen2020pedestrian} when creating the tensor of surrounding agents for each pedestrian in each video frame. The third channel of the defined tensor has a dimension equal to the number of object categories. Therefore, the position of other pedestrians and vehicles is specified separately in their corresponding layers in the channel. In \cite{zuo2021map}, two homogeneous pooling modules are defined, one for pedestrians and one for vehicles, and the heterogeneous interaction between a pedestrian and a vehicle is captured in a heterogeneous pooling module.

Another approach followed in the literature is using the category of each agent (e.g., pedestrian or vehicle) as an input feature, similar to how temporal position coordinates are encoded for each individual agent \cite{lai2020trajectory,cheng2020mcenet}. In this way, the hidden state generated for each agent will embed the agent's type, which will be implicitly accounted for when pooling the neighbouring agents together.
The type of the agent is specified using a one-hot representation in \cite{cheng2020mcenet} and is concatenated with the trajectory representation of the agent. Moreover, in this work, a separate heat maps is also created for the trajectory distribution of each agent type in order to capture the information of more likely visited areas in the past of each agent category. In \cite{chandra2019robusttp}, agent size is used as an input to account for different agent types. This size will have an effect on the weighting learned for the different interactions of a pedestrian in the attention mechanism.

In the collision probability map used in \cite{cheng2018modeling} for encoding interactions, the only difference considered for various interaction types is the way the distance between two pedestrians is calculated differently compared to the distance between a pedestrian and a vehicle in the formulation of collision probability. This different distance calculation is due to the size of the car and the elliptical shape used for modelling the vehicle in this paper.

\subsection{Interaction of Agents in Models that use Graph Neural Networks}

Among the papers that implement the structure of a graph neural network (GNN) for encoding interaction, there are a few papers that account for the difference between interactions of homogeneous agents versus interactions of heterogeneous agents~\cite{men2022pytorch,rainbow2021semantics,li2021interactive,ma2019trafficpredict,li2021hierarchical,zhang2021probabilistic,carrasco2021scout,su2022trajectory}. 

In some models proposed using graph convolutional network \cite{men2022pytorch,rainbow2021semantics}, this is done by giving labels to the graph nodes according to the agent's type and by defining a label-based adjacency matrix which specifies whether the two agents connected in the graph are of the same type or not. This label-based adjacency matrix is then used along with the velocity-based adjacency matrix for each node to predict the agent's future trajectory as a function of its interactions with others \cite{rainbow2021semantics}. 

Others are using type of agent as part of the feature vector of each node \cite{li2021interactive,ma2019trafficpredict,li2021hierarchical,zhang2021probabilistic,carrasco2021scout}. This way, the interaction type --- which is encoded uniquely based on the type of the agents interacting with each other --- can become part of the edge feature and therefore is accounted for during the prediction of agents future trajectories. 
For modelling heterogeneous traffic agents in a GNN model, a couple of papers have used a combination of an instant layer and a category layer~\cite{li2021interactive,ma2019trafficpredict,li2021hierarchical}. In the instant layer, the movement patterns of each agent in the traffic are captured. In the category layer, a super node is defined for each category of agent (e.g., pedestrian, vehicle) and each node in the instant layer is then connected to its own super node according to the agent's type. This will ensure that only agents of the same type will share the same parameters due to their similar dynamics and motion pattern.
In \cite{zhang2021probabilistic}, the same weight is used for agents of the same type when aggregating the spatial edge hidden states of all neighbouring agents.

The asymmetric interaction between multi-type agents is implicitly accounted for in the directed rate graph proposed in \cite{su2022trajectory}, where faster moving vehicles will have a different effect compared to other agents. The other models that predict trajectories of heterogeneous agents using GNN rely on using separate trajectory history encoders for each type of agent, which means that the weights of encoders are only shared among the same agent types \cite{hu2020collaborative,mo2022multi,li2021spatio}. However, these papers do not clearly distinguish between the different interaction types that a single agent can have (e.g., pedestrian-pedestrian versus pedestrian-vehicle interaction).

\subsection{Interaction of Agents in Other Methods}

Among the other methods used for modelling interaction are two papers using Cellular Automate. In these papers the interaction between pedestrian and vehicle is governed by the gap acceptance of the pedestrian, while the interaction between two pedestrians is only modelled through the restriction of not occupying the same cell at the same time \cite{cheng2019study,li2015studies}. In the method proposed in \cite{jan2020self} where pedestrians avoid vehicles by adjusting their speed and heading, the interactions among pedestrians are only discussed as their behaviour in a group. Kabtoul et al. (2020) have used different parameters in the formulation of their pedestrian motion model, one accounting for the effect of surrounding pedestrians and the other for modelling the effect of the vehicle \cite{kabtoul2020towards}. In their proposed model, the density of the space surrounding the pedestrian also impacts the pedestrian's level of cooperation with the vehicle.

\section{RQ4: Considering the uncertainties of pedestrian's predicted trajectory} \label{Q4}

\begin{figure}[b!]
  \centering
  \includegraphics[width=0.9\linewidth]{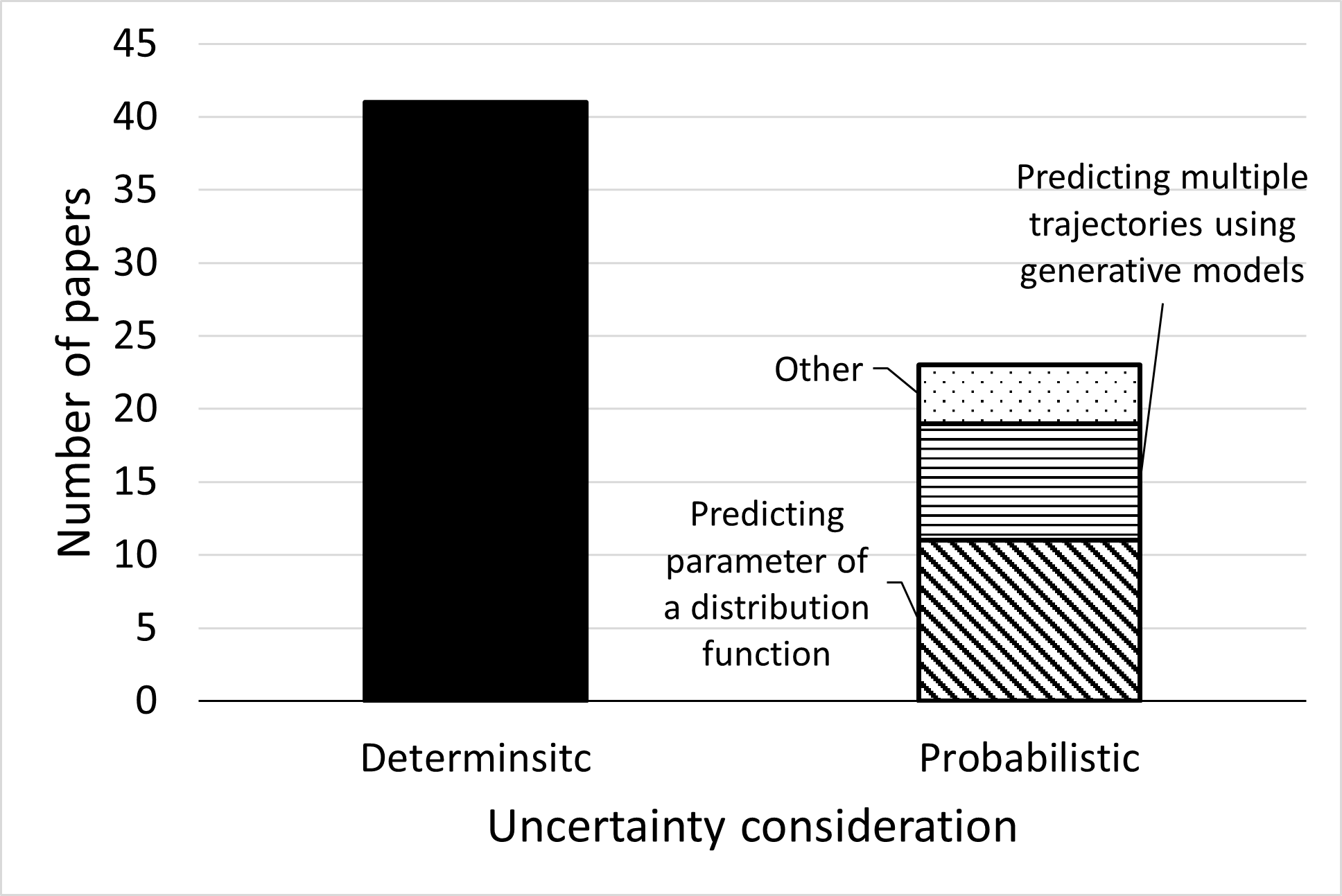}
  \caption{The number of papers using deterministic vs. probabilistic models for pedestrian motion prediction and the different methods used for considering uncertainty in the predicted trajectory.}
  \label{fig:Uncer}
\end{figure}

In reality, there are uncertainties assigned to any predicted future trajectory of pedestrians since these predictions are based on limited observable information from each person. Therefore, it is more realistic for each predicted future position to be assigned with a probability. While some methods such as the social force model simulate pedestrians' trajectories in a deterministic way, there are algorithms proposed in the literature that take into account the probabilistic nature of the pedestrian's future trajectory. By reviewing the papers, we have found three main themes of how uncertainty was modelled in the pedestrian trajectory predictions methods: 1) methods predicting the parameters of a distribution function over the future states of the pedestrian, 2) methods implementing generative models to produce multiple trajectories for each pedestrian, and 3) Other diverse methods. The number of reviewed papers in each category is shown in Fig. \ref{fig:Uncer}

\subsection{Predicting the Parameters of a Distribution Function over Pedestrians' Future Positions}

Many data-driven methods have been proposed for predicting the probability distribution of the pedestrian's next state \cite{li2021interactive,cheng2018modeling,men2022pytorch,bi2019joint,rainbow2021semantics,herman2021pedestrian,eiffert2020probabilistic,su2022trajectory}. In these models, instead of outputting the exact values of the pedestrian's next position, the network is trained to output the parameters of a bi-variate Gaussian distribution over the x and y position coordinate of the pedestrian's future trajectory~\cite{li2021interactive,cheng2018modeling,men2022pytorch,bi2019joint,rainbow2021semantics,ma2019trafficpredict,li2021hierarchical,chandra2019traphic}. Therefore, the output of the final layer of their network has five components consisting of two means, two standard deviations, and a correlation value ($\sigma_x, \sigma_y, s_x, s_y, \rho$). Some other models have used higher dimensions for the Gaussian distribution~\cite{herman2021pedestrian}. Li et al. (2021) predict both pedestrians' and vehicles' trajectories, taking into account the kinematic constraint of a vehicle's motion~\cite{li2021spatio}. Therefore, as part of their network, they predict the vehicle's next input action as a multi-variant Gaussian distribution and use the vehicle's kinematic formulation to output the vehicle's next position given its current position and the predicted action distribution~\cite{li2021spatio}.

Pedestrians' future positions are predicted as a Gaussian mixture model (GMM) in~\cite{eiffert2020probabilistic}. In this mixed model, an extra parameter representing the weighting of each Gaussian component is also added to the output of the network besides the parameters of the Gaussian distributions themselves. Instead of Gaussian distribution, Su et al. (2022), proposed the use of a Cauchy distribution for the predicted position of pedestrians. As a result, they predict the location and scale parameters of a Cauchy distribution ($m_x,m_y,\gamma_x, \gamma_y$)~\cite{su2022trajectory}.

\subsection{Using Generative Models to Produce Multiple Trajectories}

Using samples from the Gaussian distribution could cause difficulty in training by making the back propagation process non-differentiable \cite{gupta2018social}. Therefore, there is another approach followed in the literature that directly predicts the position coordinates but accounts for the uncertainty of pedestrian predicted trajectory by producing multiple trajectory outputs instead of one. This is argued to consider the multi-modality nature of the pedestrians' trajectories which means that there exist multiple acceptable future trajectories for a pedestrian given its previous trajectory~\cite{wang2021multi,zuo2021map,zhang2021probabilistic,bhattacharyya2021euro}. This is done in the related literature by using generative models either a variational auto-encoder (VAE) \cite{wang2021ltn,zhang2021probabilistic,bhattacharyya2021euro} or an Adversarial Generative Network (GAN)~\cite{zuo2021map,wang2021multi,lai2020trajectory}.

In the use of conditional variational auto-encoders (CVAE), each agent's future position is predicted given a history of its trajectory and a sample from a latent variable \cite{wang2021ltn,zhang2021probabilistic,bhattacharyya2021euro}. This latent variable itself is learned during the training phase by taking as input both the pedestrian's previous and future trajectories. When using the trained network for prediction, multiple samples from the latent variable will let producing multiple outputs for the trajectory of the pedestrian. 

The ability of Adversarial Generative Networks (GAN) to produce multiple outputs is leveraged in~\cite{zuo2021map,wang2021multi,lai2020trajectory} as another way of satisfying the multi-modality nature of pedestrians' future trajectories. This is done in \cite{zuo2021map,wang2021multi,lai2020trajectory} by inputting a random noise, sampled from a Gaussian distribution, along with the other agent's encoded features to the generator module of the GAN. These different generated paths will then be evaluated in the discriminator of the GAN structure. In the discriminator module of the GAN in~\cite{cheng2020mcenet}, a bi-variate Gaussian distribution is fit to these different samples produced by the generator and a joint probability distribution is created for storing these trajectories. The final predicted trajectory is then proposed to be the most likely prediction of the joint probability density function.

Each of these generative models have their own strength and weaknesses. GAN's are usually prone to the mode collapse problem where the network only generates one mode of trajectory with high variance, which does not thoroughly satisfy the multimodality criteria. CVAE models do not suffer from this issue by explicitly modelling multimodality, yet they present a challenge in defining additional loss criteria for proper training \cite{zhou2022dynamic}.

Therefore, for ensuring multimodality of the predicted trajectories, some models have combined the strength of both methods by proposing a hybrid CVAE-GAN model which adds the adversarial training benefit of GANs to the CVAE models \cite{zhou2022dynamic,li2019conditional}. By replacing the GAN's generator module with a CVAE structure, the hybrid model is encouraged to learn a latent space that cannot be discriminated from the latent space of the real trajectories and the overall model can produce more diverse and realistic trajectories \cite{zhou2022dynamic, li2019conditional}.

\subsection{Other Methods for Modelling Uncertainty}

We found a few other proposed methods in the literature that account for uncertainties of pedestrians' future positions~\cite{li2015studies,anderson2020off,sun2019interactive,li2021rain}. In the Cellular Automate framework used in the \cite{li2015studies}, a probability is assigned to each possible reaction of the pedestrian towards a vehicle that suddenly ran into the pedestrian critical gap acceptance. Four reactions are considered for this model, namely going forward, rushing forward, waiting, or stepping backward~\cite{li2015studies}. In \cite{anderson2020off} the next position of the pedestrian is predicted as a probability distribution over the current position and speed of the person and some other parameters such as the risk of collision with a nearby vehicle. Through the Bayesian Network framework used in~\cite{sun2019interactive}, the probability of the selected action is calculated given the current position of the agents and its discrete latent variable. By sampling an action from this distribution, the current state of the agent can be propagated to the next time step. Other methods such as adding noise to the predicted state of the agent are among the simple ways of accounting for the uncertainty of the pedestrian's predicted state~\cite{li2021rain}.

\section{RQ5: Diversity in pedestrians' behaviours} \label{Q5}

Pedestrians have different motion styles affected by their personality, age, gender, or their awareness and perception of the surrounding environment. For example, Robicquet et al. (2016) showed the existence of four navigation styles in collision avoidance strategies of pedestrians based on the distance a pedestrian keeps from others and also the distance from which the person starts the path deviation to avoid the collision \cite{robicquet2016learning}. These could be indicators of the pedestrian's level of aggressiveness in navigation that can play an important role when predicting the pedestrian's motion \cite{robicquet2016learning}.

Many models proposed for pedestrian trajectory prediction disregard these personal differences and propose an average behaviour model applied to all individuals (e.g. SFM and deep learning methods with deterministic predictions). Being able to output different trajectory styles for pedestrians can be considered a strength of a proposed model. This can be partially addressed by accounting for the uncertainty in the prediction output such as modelling the next states of a pedestrian as a probability distribution (e.g., \cite{li2021interactive,cheng2018modeling}) or producing multiple trajectories like papers that are implementing the VAE or GAN architecture (e.g., \cite{wang2021ltn,lai2020trajectory}). 

Some models have more clearly accounted for these differences in pedestrian motion behaviours. For example, two different crossing behaviours in front of a vehicle were detected in the virtual reality study conducted in \cite{kim2020pedestrian}, namely fast crossing and slow crossing. Therefore, this work has considered two separate modules for each type of pedestrian in the neural network model that captures predictions from the pedestrian perspective. Different aggressive behaviour levels in pedestrians when crossing in front of a vehicle are also modelled in \cite{li2015studies} through a parameter in the gap acceptance formulation. With this model, more aggressive pedestrians will accept a lower gap between themselves and the vehicle when crossing in front of a vehicle.

The differences between pedestrians can also appear in the cooperation level of a pedestrian when interacting with others. This level of cooperation for each pedestrian is predicted in \cite{kabtoul2020towards} as an effective factor in the pedestrian's trajectory planning model. The force-based model proposed in \cite{hesham2021advanced} simulates non-cooperative pedestrians by applying an attraction force to their current position which models a group of pedestrians that are standing still in a flow of pedestrians and blocking a passage. An irrational childlike character is modelled in \cite{jan2020self} by ignoring the vehicle's presence during the trajectory generation of those pedestrians even if the path of the vehicle is going to cross the pedestrian's path. So the pedestrian does not wait for the vehicle \cite{jan2020self}. 

Another way of considering differences between pedestrians, which is usually applied in social force models, is sampling the desired speed of each pedestrian in the simulation from a normal distribution \cite{predhumeau2022agent,predhumeau2021agent}.

\section{RQ6: Used datasets} \label{Q6}

A list of different datasets used in the papers reviewed in this article is provided in Table \ref{tab:dataset}. These datasets are categorized based on their captured environment being a conventional road, a shared space, or a campus environment. These datasets have been captured either from a bird's-eye view or from the view of a moving camera usually mounted on a vehicle.
The majority of these datasets are publicly available online.

Data-driven models for pedestrian trajectory prediction need to be trained on datasets that include trajectory data points of both pedestrians and vehicles over time. Even in some papers using physics-based methods such as the social force model, these datasets are used for calibrating the parameters of the model~\cite{anvari2015modelling,predhumeau2022agent,johora2021transferability,yang2020social} or for evaluating the performance of the algorithm by comparing the simulated trajectories with the real trajectories in the dataset~\cite{hossain2020conceptual,johora2018modeling,johora2020zone,predhumeau2021agent}. 

There already exist a variety of datasets that include trajectory data of both pedestrians and vehicles interacting with each other (e.g., DUT, INTERACTION, Waymo, nuScenes, KITTI, USyd). However, many of these datasets are captured from regular road environments (e.g., INTERACTION, Waymo, nuScenes, KITTI). Therefore, fewer datasets exist that have been recorded from less regulated environments where more diverse interaction scenarios could happen between pedestrians and vehicles, such as shared spaces or off-road campus environments (e.g., HBS, DUT, CITR, parts of SDD and USyd).

In some datasets captured by a sensor-equipped vehicle driving in the environment, annotated agents are specified in both the camera image and the LiDAR point cloud (e.g., LOKI, Argoverse, Waymo, nuScenes, KITTI; see Table~\ref{tab:dataset}). Therefore, by linking the outputs of these two sensors along with the vehicle's localization data, the trajectory of annotated agents can be visualized and used from a top-down bird's-eye view in a global frame. For example, the data points of the Waymo dataset are offered in world coordinates although the dataset was captured with a sensor-mounted vehicle \cite{llc2019waymo}.

However, there are few available datasets captured with a sensor-equipped vehicle that do not contain LiDAR point cloud data such as the PIE and JAAD datasets. In these datasets, only the ego-centric view of the dataset can be used. Therefore, in models that use these datasets (e.g., \cite{rasouli2021bifold, rasouli2019pie}), the position of the pedestrian's bounding box in the camera's image over time is used as observation and prediction sequences for the models. In the PIE dataset, the sensor readings for the ego vehicle's motion are also recorded, which is used as an input in papers focused on studying the interaction between pedestrians and the ego-vehicle~\cite{rasouli2021bifold,rasouli2019pie,quan2021holistic,yin2021multimodal,kim2020pedestrian}. On the other hand, JAAD dataset does not contain the ego vehicle's motion and instead of that specifies the high-level actions of the driver such as moving slow or speeding up.

Other datasets that are captured from a bird's-eye view, provide trajectories of all the agents in the same coordinate (e.g., SDD, HBS, HC, DUT). As these datasets are all capturing the trajectories of heterogeneous interacting agents, the different classes of agents annotated in each dataset are also provided in Table~\ref{tab:dataset}.

\noindent
\begin{table*}
\centering
\footnotesize
\caption{Datasets used for training/calibrating/evaluating pedestrian motion models. These datasets include trajectory data of heterogeneous agents including pedestrians and vehicles. Datasets specified as NM (Not Mentioned) are not given specific names by their collected teams/authors}\label{tab:dataset}
\begin{tabular}{|p{2cm}|p{2cm}|p{2cm}|p{4.5cm}|p{4.5cm}|}
\hline
\textbf{Dataset name} & \textbf{Papers using this dataset} & \textbf{Perspective} & \textbf{Captured environment} &\textbf{Annomtated heterogeneous agent classes}\\
\hline
SDD \cite{robicquet2016learning} & \cite{wang2021ltn,men2022pytorch,li2021spatio,rainbow2021semantics,wang2021multi,su2022trajectory,lai2020trajectory,cheng2020mcenet} & Bird's-eye view & Stanford University Campus environment. Both road and off-road environments & Pedestrians, Cars, Bikes, Golf carts, Skateboards \\
\hline
HBS \cite{pascucci2017discrete} & \cite{cheng2018modeling,cheng2020mcenet,johora2021transferability,johora2020agent,johora2018modeling} & Bird's-eye view & A Shared space street in Germany & Pedestrians, Cars, Cyclists\\
\hline
HC \cite{cheng2019pedestrian} & \cite{cheng2020mcenet} & Bird's-eye view & A Shared street in a university campus in Germany & Pedestrians, Vehicles, Cyclists \\
\hline
DUT \cite{yang2019top} & \cite{eiffert2020probabilistic,anderson2020off,johora2021transferability,johora2020agent,predhumeau2021agent} & Bird's-eye view & Shared space near a roundabout and an intersection both part of a Campus environment in China & Pedestrians, Cars\\
\hline
CITR \cite{yang2019top} & \cite{kabtoul2020towards,predhumeau2022agent,yang2020social,predhumeau2021agent} & Bird's-eye view & Controlled experiment in an open space area in a parking lot at Ohio State University, USA & Pedestrians and a golf cart\\
\hline
Nantes & \cite{predhumeau2022agent} & Ego-centric view & Controlled experiment in a parking lot at Ecole Centrale de Nantes, France & Pedestrians and a car\\
\hline
inD \cite{bock2020ind} & \cite{carrasco2021scout,anderson2020off} & Bird's-eye view & Four intersections in Germany & Pedestrians, Cars, Bicycles\\
\hline
INTERACTION \cite{zhan2019interaction} & \cite{mo2022multi,li2021spatio} & Bird's-eye view & Urban driving environments in US, China, Germany
and Bulgaria & Vehicles, Pedestrians \\
\hline
RounD \cite{krajewski2020round} & \cite{carrasco2021scout} & Bird's-eye view & Three Roundabouts in Germany & Cars, Trucks, Vans, Trailers, Buses Pedestrians, Bicycle, Motorcycles\\
\hline
MOT16/17/20 \cite{milan2016mot16} & \cite{chen2021modeling,chen2020pedestrian} & Ego-entric / Bird's-eye view & Unconstrained environments including Pedestrian street and urban intersection & Pedestrians, Cars, Bicycles, Motorbike, Non-motorized vehicles \\
\hline
TRAF \cite{chandra2019traphic} & \cite{chandra2019robusttp,chandra2019traphic} & Ego-entric / Bird's-eye view & Urban traffic in Asian cities & Car, Bus, Truck, Rickshaw, Pedestrian, Scooter, Motorcycle, carts\\
\hline
LOKI \cite{girase2021loki} & \cite{girase2021loki} & Ego-centric view & Urban and sub-urban driving environments in Tokyo, Japan & Pedestrian, Car, Bus, Truck, Van, Motorcycles, Bicycle\\
\hline
Argoverse \cite{chang2019argoverse} & \cite{zuo2021map} & Ego-centric view &  Urban driving environments in US & Vehicles, Pedestrians, Bicycles, Mopeds, Bus, Motorcycle, Trailers, Emergency vehicles\\
\hline
Waymo \cite{llc2019waymo} & \cite{zhang2022learning,lai2020trajectory} & Ego-centric view &  Urban and suburban traffic environments in US & Vehicles, Pedestrians, Cyclist\\
\hline
nuScenes \cite{caesar2020nuscenes} & \cite{li2021rain,wang2021ltn,hu2020collaborative,bhattacharyya2021euro} & Ego-centric view & Urban and residential driving environment in Boston and Singapore & Car, Pedestrian, Bus, Motorcycle, Trailer, Truck\\
\hline
BLVD \cite{xue2019blvd} & \cite{li2021interactive,li2021hierarchical} & Ego-centric view & Urban driving environment in China & Vehicles, Pedestrian, Riders (Cyclist, Motorcycles)\\
\hline
KITTI \cite{geiger2013vision} & \cite{quan2021holistic} & Ego-centric view & Traffic scenarios in inner-cities and rural areas in Germany & Car, Pedestrian, Van, Truck, sitting person, Cyclist, Tram, Misc (e.g, Trailer, Segway) \\
\hline
USyd \cite{sk74-7419-19} & \cite{eiffert2020probabilistic} & Ego-centric view & University of Sydney campus and surroundings & Vehicle, Pedestrian, Rider\\
\hline
ApolloScape \cite{ma2019trafficpredict} & \cite{carrasco2021scout,ma2019trafficpredict} & Ego-centric view & Urban streets in Beijing, China & Vehicle, Pedestrian, Bicycle\\
\hline
Euro-PVI \cite{bhattacharyya2021euro} & \cite{bhattacharyya2021euro} & Ego-centric view & Urban driving environments in Belgium & Pedestrians, Bicyclist and the ego-vehicle motion data\\
\hline
IVBP \cite{zhang2021probabilistic} & \cite{zhang2021probabilistic} & Ego-centric view & Urban intersection environment & Cars, Buses, Bicycles, Pedestrians, Motorcycles\\
\hline
PedX \cite{kim2019pedx} & \cite{sun2019interactive} & Ego-centric view & Three urban intersection in Michigan, USA & Only pedestrians are annotated. Trajectories of vehicles are extracted from the point cloud data in \cite{sun2019interactive}\\
\hline
PIE \cite{rasouli2019pie} & \cite{rasouli2021bifold,rasouli2019pie,quan2021holistic,yin2021multimodal,kim2020pedestrian,dos2021pedestrian} & Ego-centric view & Urban driving environments in Downtown Toronto, Canada & Pedestrians and the ego-vehicle motion data\\
\hline
JAAD \cite{rasouli2017they} & \cite{rasouli2021bifold,rasouli2019pie,quan2021holistic,yin2021multimodal} & Ego-centric view & Urban driving environments in North America and Europe & Pedestrians and the high-level ego-vehicle action\\
\hline
DAIL \cite{cai2021pedestrian} & \cite{cai2021pedestrian} & Ego-centric view & Urban streets & Pedestrian and the ego-vehicle motion data\\
\hline
NM \cite{rinke2017multi} & \cite{rinke2017multi,hossain2020conceptual} & Bird's-eye view & An intersection in university district in Braunschweig, Germany with Shared space operational features & Pedestrian, Car, Cyclist, Trucks\\
\hline
NM \cite{herman2021pedestrian} & \cite{herman2021pedestrian} & Ego-centric view & Inner-city traffic in Germany including both downtown and suburban areas & Pedestrians and ego-vehicle motion data \\
\hline
NM \cite{hassan2021predicting} & \cite{hassan2021predicting} & Bird's-eye view & Parks, shopping malls, bus stands, footpaths, and different places in educational institutes & Pedestrians and other objects including vehicles \\
\hline
BJI and TJI \cite{bi2019joint} & \cite{bi2019joint} & Bird's-eye view & Intersections in Beijing and Tianjin of China & Pedestrians (Walking, Bike, Motor), Vehicles (Car, Bus, Truck) \\
\hline
NM \cite{anvari2015modelling} & \cite{anvari2015modelling} & Bird's-eye view & Shared space street in Brighton, UK (New road street) & Pedestrians, Cars \\
\hline
NM \cite{schonauer2017microscopic} & \cite{johora2020zone} & Bird's-eye view & a shared intersection area at Sonnenfelsplatz in Graz, Austria & Pedestrians, Cars \\
\hline
NM \cite{nasernejad2021modeling} & \cite{nasernejad2022multiagent,nasernejad2021modeling} & Bird's-eye view & An intersection in Shanghai, China between two roads of Wuning and Lanxi & Pedestrians, Vehicles\\
\hline
NM \cite{rehder2018pedestrian} & \cite{ridel2019understanding} & Ego-centric view & Urban and residential driving environments & Pedestrians and ego-vehicle motion data\\
\hline

\end{tabular}
\end{table*}

\section{Gaps and Suggestions for Future Directions} \label{Future}

Pedestrian motion modelling and trajectory prediction is a well-studied area in the literature with different proposed methods. In this review, we have taken a systematic approach to identify the efforts and gaps in this area with a focus on how vehicles' effect on pedestrians' future trajectory can be modelled for unstructured environments (i.e., a place without strict traffic rules and clear landmarks and road dividers ~\cite{jyothi2019driver,kerscher2018intention}). Table \ref{tab:summary} provides a detailed summary of the findings. By reviewing these papers we have identified some gaps and see potential future research directions that suggest future work that can contribute to improvement of autonomous vehicle operations in pedestrian-crowded off-road environments.

In this review, we did not attempt to evaluate the superiority of one method over the other as for such a conclusion access to all the models and performing a detailed qualitative and quantitative comparison between all the different methods on a common benchmark is required, which was beyond the scope of this review. However, we encourage future works to focus on such comparisons to specify the best modelling approaches suitable for pedestrian trajectory prediction in unstructured environments shared with vehicles.

\subsection{Interactions in Unstructured Environments}

The focus of this paper was on pedestrian trajectory prediction methods that do not restrict their interaction scenarios to only pedestrian road crossing or using rules or geometrical information of structured environments, thus can be used for diverse interactive scenarios happening between pedestrians and vehicles in unstructured environments.

Therefore, we reviewed the existing articles that analyze pedestrians' motion in environments that are shared with vehicles such as urban shared spaces, parking lots, and off-road campus environments. We found that the majority of the methods proposed in the literature for analyzing trajectories in these environments are based on social force methods with fewer data-driven methods proposed for simulating motions in these types of environments. A probable reason can be the limited number of datasets available for shared or unstructured environments (e.g., HBS, DUT, HC, CITR, Nantes, parts of the SDD and USyd dataset) compared to datasets from road environments (e.g., inD, INTERACTION, Argoverse, Waymo, nuScense, KITTI, ApolloScape, PIE, JAAD). Due to this fact, a considerable number of deep learning-based papers reviewed here are using datasets of structured environments for model training, but they do not restrict their interaction scenarios to road crossing. Therefore, the same model architecture can be used for capturing motion behaviours in unstructured environments.

The more diverse and complex interaction that can happen between pedestrians and vehicles in unstructured environments requires more attention and is not thoroughly captured in the existing datasets. This research gap needs to be addressed by collecting more datasets from these environments and studying pedestrian motion behaviours and the vehicles' effect on their movement patterns in such spaces (e.g., at airports, shopping centres, etc.).

\subsection{Combination of Data-driven and Expert-based Methods}

Comparing data-driven methods proposed in the trajectory prediction literature with expert-based methods, data-driven methods are gaining high prediction accuracy while their black box nature sacrifices their interpretability. 
On the other hand, the outputs of expert-based motion models are explainable but they cannot always capture the complexity of human decision models. Therefore, there is more need for developing data-driven methods that can, to some extent, be explainable, such as methods that use a combination of these two types of models, as an example the one proposed in \cite{johora2020agent}, which can help overcome the drawbacks of each method and provide backup support. This combination can also help with introducing motion constraints in the deep-learning methods caused by the agent's dynamics (e.g., \cite{li2021rain,li2021spatio}) for guiding the optimization algorithm of these data-driven methods and speeding up their learning process. 

\subsection{Features Used for Modelling Interactions}

Despite the many different approaches followed in the data-driven methods for implicitly modelling interactions between pedestrians and vehicles, improvements still can be made in the features used for capturing these interactions and how two interacting agents can be specified more effectively.

The common feature used for encoding interaction in deep learning methods is the relative distance between the agents. While the trajectories of close agents are more probable to be affected by each other, distance is not the only factor governing the interactions. The impact of trajectories on one another also highly depends on the speed and the approach direction \cite{huber2014adjustments,vemula2018social}, or in other words on the risk of penetration into an agent's personal zone. Therefore, a distant agent approaching a pedestrian from the front can have a higher impact on the pedestrian's movement decision compared to a closer agent with a non-crossing path. Thus, more effective interaction features should be used for data-driven methods besides relative distance such as the relative velocity \cite{mo2022multi,li2021spatio,girase2021loki} or the collision probability \cite{cheng2018modeling} used in a few papers. These features are used to better capture the effect of agents' motion on each other, and to develop models that can better represent agents' interaction, therefore improving pedestrian prediction algorithms.

Further, almost all reviewed papers have only used the current state of the nearby agents, including the vehicle, to predict how the trajectories will interact and the way the pedestrian's trajectory will evolve in the future. However, for being able to use the pedestrian's trajectory prediction in the decision-making algorithm of an autonomous vehicle, the vehicle needs to predict the pedestrian's trajectory more actively as a function of the different actions the vehicle can take in the future \cite{camara2020pedestrian}. For example, predicting the pedestrian's response to different possible vehicle actions such as decelerating, accelerating or deviating to either side. This is related to the integration of trajectory prediction with the motion planning and control methods which is usually left unexplained in papers that focus on either of these areas \cite{rudenko2020human}. This level of prediction capability that accounts for the vehicle's future action in the pedestrians' following trajectory is necessary, especially for simulation environments in which a reinforcement learning navigation algorithm is trained for vehicle operation among pedestrians. This will let the vehicle's navigation algorithm take an active role in controlling the interaction between the pedestrian and the vehicle based on these predictions. This is especially an important consideration for unstructured environments where the vehicle needs to be trained on more diverse interaction scenarios which have been studied less as the majority of the previous research and their collected dataset are based on structured environments.

These more active pedestrian trajectory prediction algorithms, if data-driven, will also require a more detailed dataset that contains the vehicle's action at each step when interacting with the pedestrian. Therefore, more detailed future datasets containing the vehicle's action on the brake and throttle as well as the steering wheel can benefit this line of research.

\subsection{Considering Pedestrians' Different Motion Styles}

Most papers in the current literature assume the same behaviour for all pedestrians in the environment, predicting all the motions from the same model~\cite{rudenko2020human}. Only \cite{kim2020pedestrian,li2015studies} considered pedestrians' different crossing styles and \cite{kabtoul2020towards} formulated the cooperation factor of pedestrians based on other parameters, with non-cooperative or irrational pedestrians behaviours being simulated in \cite{hesham2021advanced,jan2020self}. Therefore, based on the findings of section \ref{Q5}, pedestrian differences in how they navigate and cooperate with others for collision avoidance, while being important, has seen limited attention in the literature. How pedestrians' reactions toward a vehicle can vary based on the pedestrians' time pressure, age or the person's overall navigation style can highly affect the future prediction of a pedestrian's trajectory. Therefore, there is a need for more comprehensive pedestrian motion models that can encode these differences. This is especially important to be considered in data-driven methods since they are prone to learn an average behaviour seen in the dataset unless trained in a way to explicitly account for the different navigation behaviours embedded in the data. An intelligent autonomous vehicle that can perform at a social awareness level same as a human driver needs to recognize these differences in the way pedestrians handle interactions.

\subsection{Benchmarks for Model Evaluation in Unstructured Shared Spaces}

The performance of each proposed pedestrian trajectory prediction model in the literature is usually reported in comparison to a few selected baselines over some subsets of the existing trajectory datasets. This prevents one from being able to conduct a generalizable and fair quantitative comparison among all different approaches. For making this comparison possible, a common benchmark is required that tests all the algorithms on the exact same datasets and ideally can use richer evaluation metrics that can also measure social awareness of the predicted trajectories, specially for those models that encode interactions effects. Such a comparison could give the community insights on model architectures and individual module designs that perform better than others and suggest areas for potential improvement. In addition, effective learning and evaluation of different interaction modules require training and testing these models on highly interactive scenarios. In fact, having many non-interactive trajectories in a dataset can prevent the model from truly focusing on learning interaction patterns and can lead to misleading evaluations \cite{Kothari2020HumanTF}.

In this regard, the TrajNet++ benchmarks for pedestrian trajectory prediction has been introduced in 2020 \cite{Kothari2020HumanTF}.
By defining a high level trajectory categorization consisting of \textit{Static}, \textit{Linear}, \textit{Interacting}, and \textit{Non-interacting} trajectory types, this benchmark focuses on the \textit{Interacting} category and has sampled these types of scenarios from a couple of popular pedestrian trajectory datasets for building the TrajNet benchmark. Therefore, this benchmark focuses on training and evaluating pedestrian trajectory prediction in highly interactive scenarios. This benchmark also introduces two novel metrics related to collision for better evaluating the collision avoidance capabilities of different models which is a crucial metric for evaluating the level of social-awareness of the predicted trajectories in interactive scenarios.

Kothari et al. (2020) have compared three different grid-based and four different non-grid based interaction modules (e.g., max-pooling, sum-pooling, attention and concatenation) on this benchmark while keeping the remaining modules (encoder and decoder) the same \cite{Kothari2020HumanTF}. Their results suggest that among non-grid based methods in interaction modelling, using attention mechanisms for aggregating the trajectory information of neighbouring agents gives lower prediction errors and specially lower collision rates compared to using other aggregation models such as concatenation and max-pooling. Meanwhile they highlight the fact that despite its simplicity, concatenating the embedding of various neighbours for interaction modelling performs only slightly worse compared to its more complex attention-based counterpart \cite{Kothari2020HumanTF} but still works better then both max-pooling and sum-pooling by preserving the information of the surrounding neighbours. Similar compassion exists for interactions captured through a grid-based methods where their results suggest that grids built based on velocity information can more accurately predict the future trajectories of pedestrians with less collisions \cite{Kothari2020HumanTF}.

Despite all its benefits, the TrajNet benchmark is focused on pedestrians trajectories only in pedestrian-pedestrian interaction scenarios and lacks datasets that contain both pedestrians and vehicles in the environment which was the focus of the models reviewed in this paper. Comparisons on a common benchmark that includes both pedestrians' and vehicles' trajectories, though not exists yet, can be beneficial for evaluating other aspects discussed in this paper such as for specifying the most efficient method for encoding uncertainty or multimodality among the existing propose models (e.g., GAN vs CVAE).

There are other open challenges and benchmarks such as the Argoverse2 motion forecasting challenge \cite{wilson2023argoverse}, the Waymo challenge \cite{ettinger2021large} or the nuScenes prediction benchmark \cite{caesar2020nuscenes} that are named after their datasets which are collected from the road environments. These benchmarks are mainly focused on vehicle trajectory prediction and also include pedestrian trajectories on the sidewalks and crosswalks, but do not really report separate evaluations for pedestrians and vehicles predicted trajectories. Therefore, a separate benchmark or at least an extension to the existing ones is required which could evaluate the pedestrians' trajectories in scenarios where both pedestrian-pedestrian and pedestrian-vehicle interaction exist in shared and more unstructured spaces. Such a benchmark could then be used for evaluating and comparing the models that can be used for pedestrians trajectory prediction in an unstructured environment where both pedestrians and vehicles are present. However, we believe that the main barrier for such benchmarks covering heterogeneous interaction types in shared unstructured spaces is the limited number of existing datasets from such environments. Therefore, as a first step, more datasets should be collected from these environments as discussed previously, based on which benchmarks can be developed in the future.

\section{Limitations} \label{lim}

Our review has a couple of limitations which are typical for survey papers. We did our best to carefully design a comprehensive search query through an iterative process and also consulting with the Liaison Librarian at University of Waterloo. But it is always possible that we might have missed some related articles through these search queries.
Also, as commonly used in review articles related to topics addressed in this review, we selected the three popular databases known for containing papers in this research area. But there could have been articles in other databases that we have not covered. Further, our review has not included articles that were written in a language other than English.

Unstructured environment is not a common terminology used in the literature. It is very recently formally defined \cite{Golchoubian2022,jyothi2019driver,kerscher2018intention}. Thus, to be consistent in the selection process, we had to carefully define a criteria for selecting the papers that could satisfy the condition for our environmental consideration during the screening and eligibility checks. To take into account the different terminologies used in different work, we did not include any term in the search queries related to unstructured environments. We rather manually checked all the articles for the type of environment, the considered interaction scenarios, and the environmental information used in the model. We then kept those articles that used an environment that could be considered as unstructured (based on the definition provided earlier in the review and the related literature defining it) or had proposed a model that can potentially be used for unstructured environments, i.e., models that did not restrict the interaction scenario to road crossing or used any traffic rule or road-specific information in the model. 

When reviewing and summarizing the reviewed articles, we only relied on the information reported by the authors about the model. 
Considering this fact, it was not always clear if the articles using heterogeneous datasets had considered both pedestrians and vehicles in the data. Therefore, we only reported on papers that had clearly specified considerations of vehicles and pedestrians simultaneously in their model.

Finally, the list of datasets provided here is based on the ones we found in the reviewed articles, where we tried to include even less popular datasets. But there might be other datasets of unstructured environments or mixed pedestrian-vehicle spaces that we did not find.

\section{Conclusion} \label{Conc}

Models that can represent pedestrian-vehicle interactions in unstructured environments require more attention as interaction effects can be different and more diverse in these less-regulated, shared spaces compared to road environments. This becomes particularly important as low-speed autonomous vehicles in the future are visioned to give service in current pedestrian-dominated environments or share the space more commonly with pedestrians on shared streets. These will require autonomous vehicles to have a better understating of the effect of their presence and actions on pedestrians' future motion for guaranteeing pedestrian safety.

This article reviewed the existing work on pedestrian motion modelling or trajectory prediction in the presence of vehicles. It provided a summary of (1) Pedestrian motion prediction approaches, (2) the approaches used for modelling pedestrian-vehicle interaction, (3) considered differences between pedestrian-pedestrian interaction versus pedestrian-vehicle interaction (4) uncertainty modelling in the pedestrian's predicted trajectories (5) pedestrian behaviour diversity consideration, and (6) the datasets used that contains both pedestrian and vehicles in a mixed environment. Through this review research gaps were identified as below:

\begin{itemize}
    \item More comprehensive studies on diverse interactions between pedestrians and vehicles in unstructured environments are required, collecting more data from such environments.
    \item There is a need for developing models that combine data-driven and expert-based models that can bring together the benefits of both approaches (e.g, the interpretability of expert-based models and the ability of the data-driven models in capturing more complex behaviours).
    \item There is a need to study using interaction features that are beyond just relative distance between agents and take into account the active influence of vehicle's decision on the pedestrian motion in the prediction models.
    \item Pedestrians' different motion styles in the trajectory prediction need to be considered in the future work. 
    \item Different methods need to be evaluated on a common benchmark for a better understating of different methods strength and weaknesses.
\end{itemize}

These identified gaps can inspire the future research direction for pedestrian trajectory prediction under influence of vehicles in unstructured environments.

\appendices{
\begin{landscape}
\pagestyle{empty}
\begin{table}
    \caption{Summary of reviewed papers (\textit{Ped} stands for pedestrian and \textit{Veh} stands for vehicle).}
    \footnotesize
    \label{tab:summary}

    \centering
    \footnotesize
    \begin{tabularx} {23cm} {|p{2cm} | p{2cm} | p{4cm} | p{4cm} | p{4cm} |p{2cm} | p{2cm} |}
\hline
\bf Paper  & \bf Motion prediction model & \bf Interaction model & \bf Ped-Ped vs. Ped-Veh interaction & \bf Uncertainty consideration & \bf Pedestrian diversity &  \bf Environment \\
\hline

Rasouli et al. (2021) \cite{rasouli2021bifold} & Deep-learning & Pooling through an interaction attention unit and using ego vehicle's speed & Separate semantic map for each agent type & Deterministic & - & Road datasets\\
\hline
Chi Zhang et al. (2022) \cite{zhang2022learning} & Deep-learning (CNN/LSTM) & Max pooling & Separate pooling modules one for pedestrian and one for vehicle & Deterministic & - & Road dataset\\
\hline
Yang et al. (2017) \cite{yang2017agent} & Social Force Model & Repulsive forces & Different formulation for repulsive forces from ped vs. veh & Deterministic & - & Shared space\\
\hline
Hossain et al. (2020) \cite{hossain2020conceptual} & Social Force Model + Game theory & Repulsive forces and game modelling & The type of conflict being ped-ped or ped-veh is specified in their defined conflict graph & Deterministic & - & Shared space\\
\hline
Jiachen Li et al. (2021) \cite{li2021rain} & Deep learning + kinematic model & Edges in the Graph Neural Network & Not clear & Noise addition to the predicted state & - & Road dataset\\
\hline
Zirui Li et al. (2021) \cite{li2021interactive} &  Deep learning (GNN) & Edges in the GNN with relative distance as an edge feature & The type of interacting agents is also an edge feature & Predicting parameters of a Bi-variant Gaussian Distribution  & - & Road dataset\\
\hline
YingQiao Wang et. al (2021) \cite{wang2021ltn} & Deep learning & Pooling through a weighted sum (additive attention) & Not clear & Using CVAE to produce multiple trajectories & - & Campus and road dataset\\
\hline
Zuo et al. (2021) \cite{zuo2021map} & Deep learning & Max pooling & Using a homogeneous pooling module for ped-ped interaction and a heterogeneous pooling module for ped-veh interaction & The GAN structure with a white noise vector produces multiple trajectories & - & Road dataset\\
\hline
Carrasco et al. (2021) \cite{carrasco2021scout} & Deep learning (GCN) & Adjacency matrix based on edge connections in GNN & Agent type is a feature of the graph's node & Deterministic & - & Road dataset\\
\hline
Hu et al. (2020) \cite{hu2020collaborative} & Deep learning & Directed edges in the GNN with relative positions as an edge feature & Not clear & Deterministic & - & Road dataset\\
\hline
Cheng et al. (2021) \cite{cheng2018modeling} & Deep learning & Pooling using a probability density map & The different size and shape of the vehicle appears in the calculation of distance within the collision probability map & Predicting parameters of a Bi-variant Gaussian Distribution & - & Shared space\\
\hline
Rasouli et al. (2019) \cite{rasouli2019pie} & Deep learning & Using ego vehicle's speed as input & Only interaction between pedestrians and the ego-vehicle is modelled & Deterministic & - & Road dataset\\
\hline
Cai et al. (2022) \cite{cai2021pedestrian} & Deep learning - CNN encoder-decoder & Considering the effect of vehicle's ego-motion on the pedestrian through a transformation module & Only interaction between pedestrians and the ego-vehicle is modelled & Deterministic & - & Road dataset\\
\hline
Hesham et al. (2021) \cite{hesham2021advanced} & Force-based model & Repulsive force is applied when personal spaces collide & Not clear & Deterministic & Also simulating non-cooperative pedestrians & An ambulance passing through a dense crowd\\
\hline
Herman et al. (2021) \cite{herman2021pedestrian} & Deep learning & Using ego vehicle's distance to the pedestrian as input &  Only interaction between pedestrians and the ego-vehicle is modelled & A probability density function is predicted with the CVAE model & - & Road dataset\\
\hline
Yang et al. (2018) \cite{yang2018crowd} & Social Force Model & Repulsive forces &  Different formulation for repulsive forces from ped vs. veh & Deterministic & - & Shared space\\
\hline
Janapalli et al. (2019) \cite{janapalli2019heterogeneous} & Force-based & collision penalty force & Not clear & Deterministic & - & An ambulance passing through a dense crowd\\
\hline
Hassan et al. (2021) \cite{hassan2021predicting} & Deep learning & Pooling using occupancy grid map & Not clear & Deterministic & - & Roads, parks, footpaths, bust stands, etc.\\
\hline
Men et al. (2022) \cite{men2022pytorch} & Deep learning (GCN) &  Adjacency matrix based on edge connections in GNN  & A label-based adjacency matrix is defined & Predicting parameters of a Bi-variant Gaussian Distribution & - & Campus environment dataset\\
\hline
Quan et al. (2021) \cite{quan2021holistic} & Deep learning & Using ego vehicle's speed as input & Only interaction between pedestrians and the ego-vehicle is modelled & Deterministic & - & Road dataset\\
\hline

    \end{tabularx}
\end{table}
\end{landscape}


\begin{landscape}
\pagestyle{empty}
\begin{table}
     \captionsetup{labelformat=empty}
    \caption{TABLE III Continued}
    \centering
    \footnotesize
    \begin{tabularx} {23cm} {|p{2cm} | p{2cm} | p{4cm} | p{4cm} | p{4cm} |p{2cm} | p{2cm} |}
\hline
\bf Paper  & \bf Motion prediction model & \bf Interaction model & \bf Ped-Ped vs. Ped-Veh interaction & \bf Uncertainty consideration & \bf Pedestrian diversity &  \bf Environment \\
\hline
Yin et al. (2021) \cite{yin2021multimodal} & Deep learning  (Transformer) & Ego vehicle's speed sequence is used as input & Only interaction between pedestrians and the ego-vehicle is modelled & Deterministic & - & Road dataset\\
\hline
Lin Zhang et al. (2022) \cite{zhang2021pedestrian} & Constant velocity and constant acceleration & Coupling motions by transforming pedestrian coordinate to the vehicle moving coordinate & Only interaction between one pedestrian and one vehicle is modelled & Deterministic & - &  Simulated road\\
\hline
Kim et al. (2020) \cite{kim2020pedestrian} & Deep learning & Ego vehicle's speed in used as input & Only interaction between pedestrians and the ego-vehicle is modelled & Deterministic & Two models for two crossing behaviours of pedestrians (fast and slow) & Road dataset\\
\hline
Anvari et al. (2014) \cite{anvari2014long} & Social Force Model & Repulsive force & Different parameters for the repulsive forces and considering the size and shape of the vehicle & Deterministic & - & Shared space \\
\hline
Kai Chen at al. (2021) \cite{chen2021modeling} & Deep learning (ConvLSTM) & Pooling using an occupancy grid map & Different channels in the occupancy map for different agent types & Deterministic & - & Roads, Pedestrianized spaces \\
\hline
Santos et al. (2021) \cite{dos2021pedestrian} & Deep learning & Ego vehicle's speed sequence is used as input  & Only interaction between pedestrians and the ego-vehicle is modelled & Deterministic & - & Road dataset \\
\hline
Borsche et al.(2019) \cite{borsche2019microscopic} & Social Force Model & Repulsive forces & Repulsive forces are different in strength and the interaction range & Deterministic & -&  Simulated road environment \\
\hline
Jan et al. (2020) \cite{jan2020self} & Adjusting speed and heading according to some heuristics & Using distance to collision to adjust speed and heading direction & Interaction among pedestrians is only discussed in group & Deterministic & Also simulating irrational or childlike characters by ignoring the vehicle & Pedestrian zone in a campus environment \\
\hline
Eiffert et al. (2020) \cite{eiffert2020probabilistic} & Deep learning & Pooling through a weighted sum & Vehicle's effect on the ped-ped interaction is modelled & Parameters of a Gaussian Mixture Model for the trajectories are predicted  & - & Campus environment dataset\\
\hline
Mo et al. (2022) \cite{mo2022multi} & Deep learning (GNN) & Directed edges of a GNN with relative position, velocity and heading as edge features & Edge types are defined by a concatenation of the type of the two connected nodes and used as an edge feature & Deterministic & - & Road dataset \\
\hline
Bi et al. (2019) \cite{bi2019joint} & Deep learning & Pooling using an occupancy grid map & Separate occupancy maps for pedestrians and vehicles & Predicting the parameters of a Bi-variant Gaussian Distribution & - & Road dataset \\
\hline
Yang et al. (2018) \cite{yang2018social} & Social Force Model & Repulsive forces  & Different formulation for repulsive forces from ped vs. veh & Deterministic & - & Interaction scenarios in unstructured environment \\
\hline
Cheng et al. (2019) \cite{cheng2019study} & Cellular Automate + model calibration with data & The gap acceptance specify whether the pedestrian will cross in front of the vehicle & Ped-ped interaction model is limited to not occupying the same cell at once but ped-veh interaction is modelled based on gap acceptance & Deterministic & - & Drop-off area at a railway station \\
\hline
Jiachen Li et al. (2021) \cite{li2021spatio} & Combination of Deep learning and Physics-based dynamic models & Edges in GNN with relative position, velocity and heading as edge attributes & Not clear & The parameters of a multi-variant Gaussian Distribution is predicted for the control inputs & - & Campus environment and road dataset \\
\hline
Anderson et al. (2020) \cite{anderson2020off} & Adjusting velocity based on a piece-wise linear vehicle influence and a risk function & A vehicle influence function specifies the pedestrian velocity change & Only ped-veh interaction is modelled with no ped-ped interaction &  Next position is modelled as a probability of the current observed position while expanding the joint distribution model & - & Shared space \\
\hline

    \end{tabularx}
\end{table}
\end{landscape}


\begin{landscape}
\pagestyle{empty}
\begin{table}

    \captionsetup{labelformat=empty}
    \caption{TABLE III Continued}
    \centering
    \footnotesize
    \begin{tabularx} {23cm} {|p{2cm} | p{2cm} | p{4cm} | p{4cm} | p{4cm} |p{2cm} | p{2cm} |}
\hline
\bf Paper  & \bf Motion prediction model & \bf Interaction model & \bf Ped-Ped vs. Ped-Veh interaction & \bf Uncertainty consideration & \bf Pedestrian diversity &  \bf Environment \\
\hline

Rainbow et al. (2021) \cite{rainbow2021semantics} & Deep learning (GCN) &  Adjacency matrix based on edge connections in GNN  & A label-based adjacency matrix is defined & Predicting the parameters of a Bi-variant Gaussian distribution & - & Campus environment dataset\\
\hline
Yu Wang et al. (2021) \cite{wang2021multi} & Deep learning (CNN+LSTM) & Pooling using occupancy grid map & No difference made & Using GAN to generate multiple trajectories & - & Campus environment dataset\\
\hline
Ma et al. (2019) \cite{ma2019trafficpredict} & Deep learning (GNN) & Edges in GNN with relative distance as an edge feature & The type of interacting agents is also an edge feature & Predicting the parameters of a Bi-variant Gaussian distribution & - & Road dataset \\
\hline
Su et al. (2022) \cite{su2022trajectory} & Deep learning (GCN) & Adjacency matrix based on directed edge connections in GNN & Velocity difference between ped and veh is implicitly encoded in the rate graph & Predicting parameters of a Cauchy distribution & - & Campus environment dataset\\
\hline
Lai et al. (2020) \cite{lai2020trajectory} & Deep learning & Pooling through a weighted sum & The category of agents gets encoded in their temporal feature so is considered when pooling & Using the GAN structure to produce multiple trajectories & - & Campus environment and road datasets \\
\hline
Girase et al. (2021) \cite{girase2021loki} & Deep learning & Directed edges of GNN with relative position and velocity as edge attributes & Not clear & Deterministic & - & Road dataset \\
\hline
Bhattacharyya et al. (2021) \cite{bhattacharyya2021euro} & Deep learning & Interactions effect is captured in a joint latent space within the VAE architecture & Not clear & Predicting multiple trajectories using the latent variable in the VAE architecture & - & Road dataset \\
\hline
Chandra et al. (2019) \cite{chandra2019robusttp} & Deep learning (LSTM+CNN) & Pooling in a map & Using agent size as an input & Deterministic & - & Road dataset \\
\hline
Zhao Zhang et al. (2022) \cite{zhang2022modeling} & Social Force Model & Repulsive force & Different distance calculation due to vehicle's elliptical shape  & Deterministic & - & Parking lot \\
\hline
Cheng et al. (2020) \cite{cheng2020mcenet} & Deep learning (LSTM + CNN) & Pooling using a polar occupancy grid map & Type of agents is specified using a one-hot representation and is concatenated with the motion representation & Using a CVAE architecture they generate multiple trajectories and fit a Bi-variant Gaussian Distribution to it & - & Shared space and a campus environment dataset \\
\hline
Chen et al. (2021) \cite{chen2020pedestrian} & Deep learning (ConvLSTM) & Pooling using an occupancy grid map & Each agent type has a different channels in the occupancy map & Deterministic & - & Roads, Pedestrianized spaces dataset \\
\hline
Anvari et al. (2015) \cite{anvari2015modelling} & Social Force Model + model calibration with data & Repulsive force & Different parameters for the repulsive forces and considering the size and shape of the vehicle & Deterministic & - & Shared space \\
\hline
Xuexiang Zhang et al. (2021) \cite{zhang2021probabilistic} & Deep learning (GNN) & Directed edges in GNN with relative distance as an edge feature & Each agent has a semantic category and the parameters are shared between nodes and edges of with same semantic type & Using the structure of a CVAE to generate multiple trajectories & - & Road datasets\\
\hline
Johora et al. (2018) \cite{johora2018modeling} & Social Force Model + Game theory & Repulsive force and game & Repulsive forces are different in strength and the interaction range  & Deterministic & - & Shared space \\
\hline
Sun et al. (2019) \cite{sun2019interactive} & Bayesian Network & Through causal connections in the graphical model & No difference except for having different input states and motion model for veh vs. ped & By predicting a probability distribution for the selected action and sampling from that, the next state will be generated using their model & - & Road dataset \\
\hline
Nasernejad et al. (2022) \cite{nasernejad2022multiagent} & Multi agent DRL Planning method & The joint policy is solved using DRL. Interaction is also embedded in the reward function recovered with IRL & No ped-ped interaction is modelled 
& Deterministic & - & Road dataset\\
\hline
Ridel et al. (2019) \cite{ridel2019understanding} & Deep learning & Using ego vehicle’s position sequence as input & Only interaction between pedestrians and the ego-vehicle is modelled & Deterministic & - & Road datasets\\
\hline
Johora et al. (2020) \cite{johora2020zone} & Social Force Model + Game & Repulsive force and game solving & Different conflict types are solved differently and having an additional game layer for ped-veh interactions & Deterministic & - & Shared space \\
\hline

    \end{tabularx}
\end{table}
\end{landscape}


\begin{landscape}
\pagestyle{empty}
\begin{table}

    \captionsetup{labelformat=empty}
    \caption{TABLE III Continued}
    \centering
    \footnotesize
    \begin{tabularx} {23cm} {|p{2cm} | p{2cm} | p{4cm} | p{4cm} | p{4cm} |p{2cm} | p{2cm} |}
\hline
\bf Paper  & \bf Motion prediction model & \bf Interaction model & \bf Ped-Ped vs. Ped-Veh interaction & \bf Uncertainty consideration & \bf Pedestrian diversity &  \bf Environment \\
\hline
Rinke et al. (2017) \cite{rinke2017multi} & Social Force Model & Repulsive force & Not clear & Deterministic & - & Shared space \\
\hline
Li et al. (2015) \cite{li2015studies} & Cellular Automata & The gap acceptance model specifies whether the pedestrian will cross in front of the
vehicle & Ped-ped interaction model is limited to not occupying the same cell at once but ped-veh interaction is modelled based on gap acceptance & When a vehicle gets into a pedestrian critical gap a probability is assigned to possible reaction behaviours (e.g., going forward, rushing, waiting or going backward) & The aggressiveness of the pedestrian is specified as a parameter& Simulated road\\
\hline
Johora et al. (2021) \cite{johora2021transferability} & Social Force Model + Game + model calibration with data & Repulsive force and the joint optimal policy in a game & Repulsive forces are different in strength and the interaction range & Deterministic & - & Shared space \\
\hline
Johora et al. (2020) \cite{johora2020agent} & Combined expert-based and deep learning & Repulsive force and game and pooling & Different calculation of distance in pooling and different repulsive force formulation & Deterministic & - & shared space \\
\hline
Kabtoul et al. (2020) \cite{kabtoul2020towards} & Linear regression model & The probability of collision and the deformation of the personal zone by the vehicle are among influencing factors on the pedestrians predicted speed  & The effect of surrounding pedestrians is defined as a separate parameter from vehicle's influence & Deterministic & Different cooperation factors are considered for pedestrians & Unstructured environment, Parking lot dataset\\
\hline
Zirui Li et al. (2021) \cite{li2021hierarchical} & Deep learning (GNN) & Edges of GNN with the relative position as an edge feature & The type of interacting agents is also an edge feature & Predicting parameters of a Bi-variant Gaussian Distribution & - & Road dataset \\
\hline
Predhumeau et al. (2022) \cite{predhumeau2022agent} & Social Force Model + Decision model + model calibration with data & Repulsive force & Different parameters in the repulsive forces formulation and some additional forces for ped-veh interaction & Deterministic & Preferred walking speed was sampled from a normal distribution & Shared space \\
\hline
Yang et al. (2020) \cite{yang2020social} & Social Force Model + model calibration with data & Repulsive force & Vehicle's orientation, size, shape and velocity play a role in the repulsive force & Deterministic & - & Shared space \\
\hline
Nasernejad et al. (2021) \cite{nasernejad2021modeling} & Deep Reinforcement learning & The relative distance, speed and heading angle are parameters used in the model. Interaction is also embedded in the reward function recovered with IRL & No ped-ped interaction is modelled & Deterministic & - & Road dataset \\
\hline
Predhumeau et al. (2021) \cite{predhumeau2021agent} & Social Force Model + Decision model & Repulsive force & Different parameters in the repulsive forces formulation and some additional forces for ped-veh interaction & Deterministic & Preferred walking speed was sampled from a normal distribution & Shared space \\
\hline
Chandra et al. (2019) \cite{chandra2019traphic} & Deep learning (LSTM + CNN) & Pooling in a constructed neighbour and horizon map & The agent's size is an input to the model. The weighting of different interactions is learned based on shape, dynamic constraints and behaviour of agents & Predicting the parameters of a Bi-variant Gaussian distribution & - & Road dataset \\
\hline

    \end{tabularx}
\end{table}
\end{landscape}
}

\section*{Acknowledgments}
This research was undertaken, in part, thanks to funding from the Canada 150 Research Chairs Program and the Natural Sciences and Engineering Research Council of Canada (NSERC). We would like to thank Rebecca Hutchinson, the Liaison Librarian for Computer Science at University of Waterloo for helping with design of the search query of this systematic review.



\bibliographystyle{IEEEtran}
\bibliography{IEEEabrv,mybibfile}

\newpage

\section{Biography Section}
 





\vspace{-33pt}

\begin{IEEEbiography}   [{\includegraphics[width=1in,height=1.25in,clip,keepaspectratio]{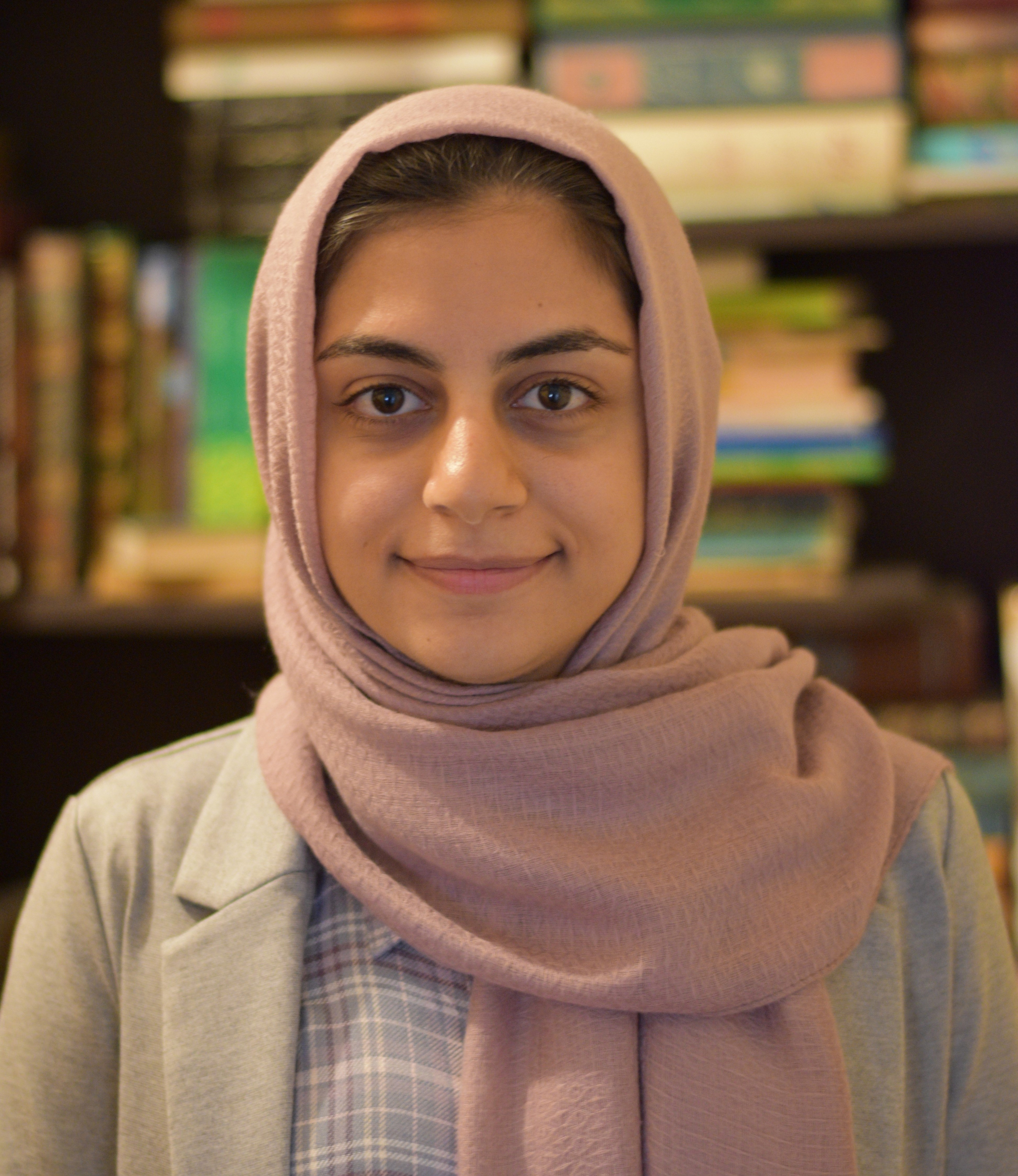}}]{Mahsa Golchoubian} is a Ph.D. candidate at the University of Waterloo’s Systems Design Engineering Department, Canada. She received her B.Sc. and M.Sc. in Aerospace Engineering from Sharif University of Technology, Iran, in 2015 and 2018, respectively. Currently, she is conducting research at the intersection of autonomous navigation, human-robot interaction, and machine learning.
\end{IEEEbiography}

\begin{IEEEbiography} 
[{\includegraphics[width=1in,height=1.25in,clip,keepaspectratio]{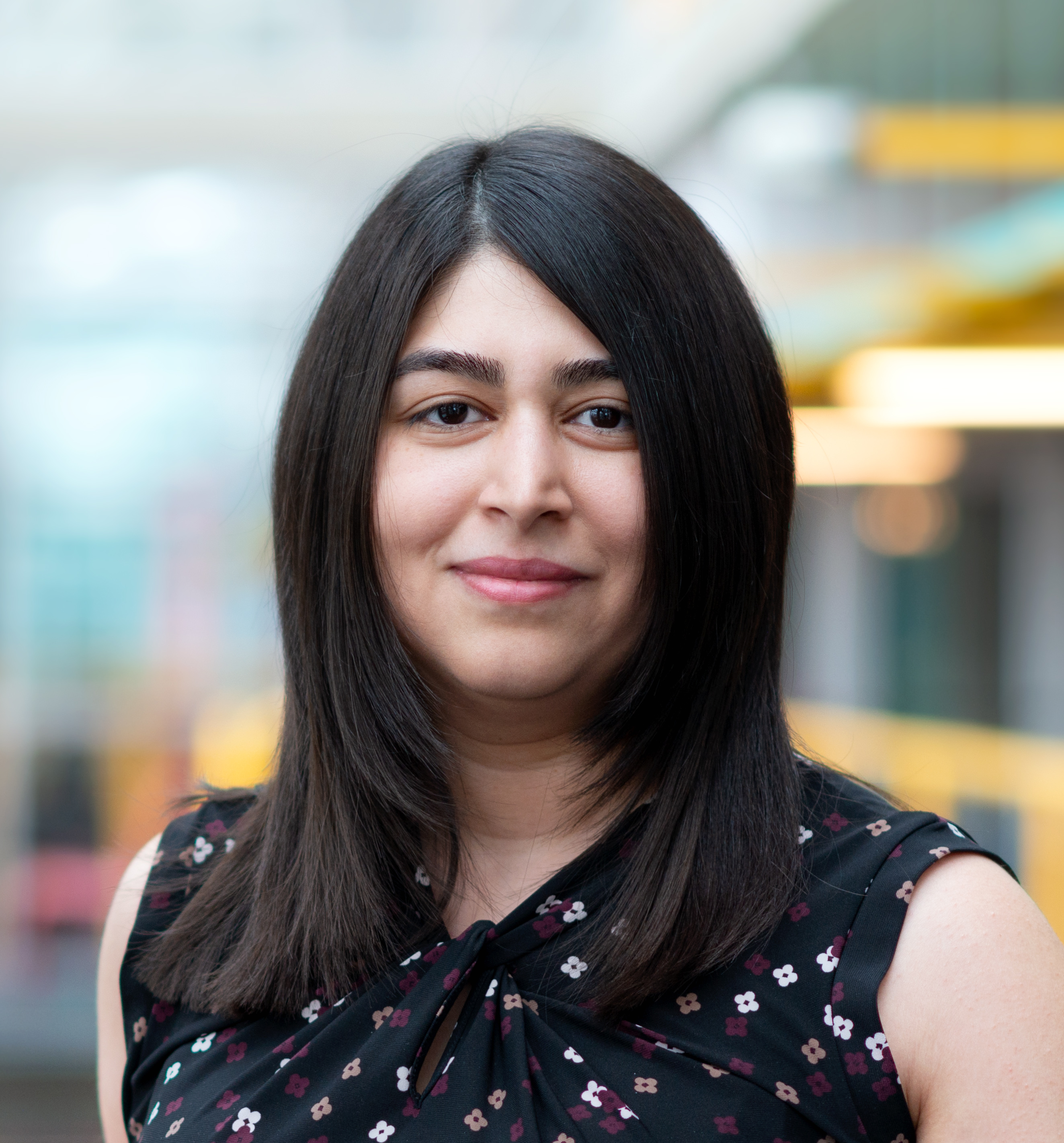}}]{Moojan Ghafurian} is a Research Assistant Professor in the Department of Electrical and Computer Engineering at the University of Waterloo. She got her PhD from the Pennsylvania State University and was the Inaugural Wes Graham postdoctoral fellow from 2018-2020 at David R. Cheriton School of Computer Science at the University of Waterloo. Her research areas are human-computer/robot interaction, social robotics, affective computing, and cognitive science. Her research explores computational models of how humans interact with systems to inform user-centered design of emotionally intelligent agents in multiple domains.
\end{IEEEbiography}

\begin{IEEEbiography} 
[{\includegraphics[width=1in,height=1.25in,clip,keepaspectratio]{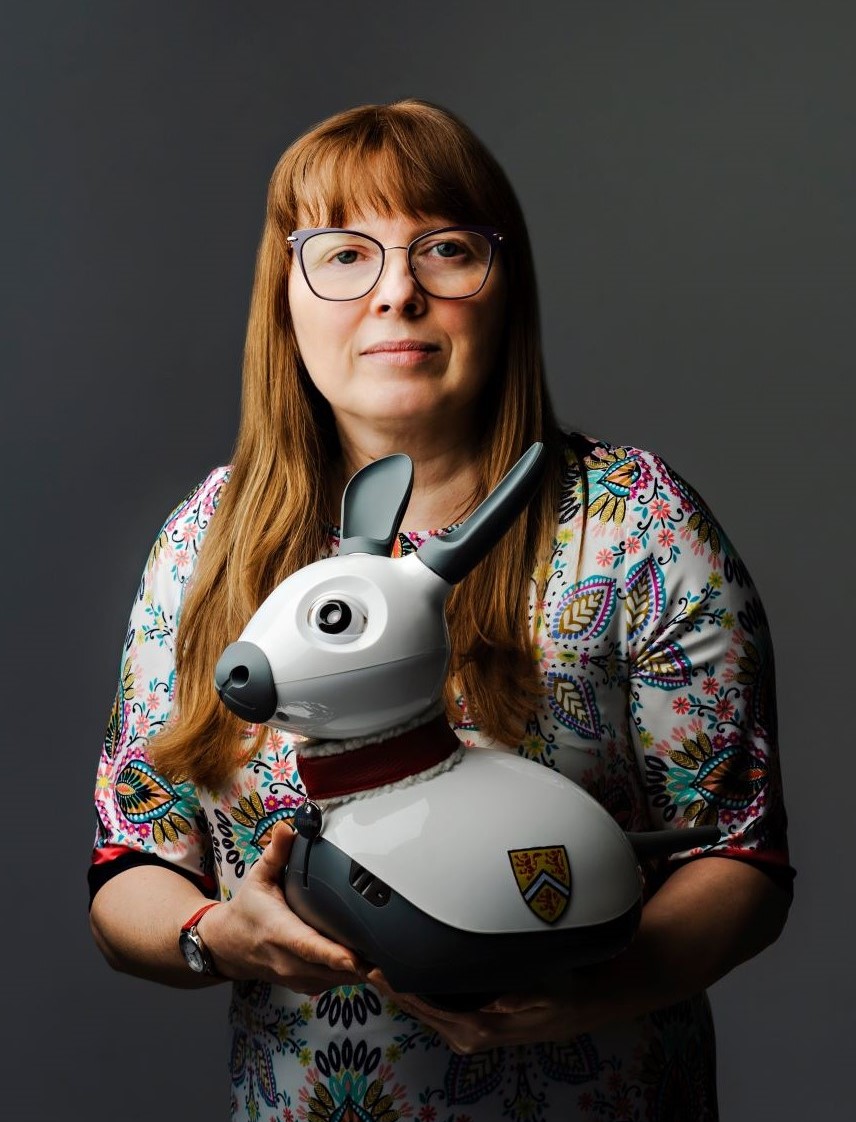}}]{Kerstin Dautenhahn} is Canada 150 Research Chair in Intelligent Robotics, Faculty of Engineering, University of Waterloo, Canada where she directs the Social and Intelligent Robotics Research Laboratory (SIRRL). She became IEEE Fellow for her contributions to Social Robotics and Human-Robot Interaction. From 2000-2018 she coordinated the Adaptive Systems Research Group at University of Hertfordshire, UK. Her main research areas are human-robot interaction, social robotics, cognitive and developmental robotics, and assistive technology, with applications of social robots as tools in education, therapy, healthcare and wellbeing.
\end{IEEEbiography}

\begin{IEEEbiography} 
[{\includegraphics[width=1in,height=1.25in,clip,keepaspectratio]{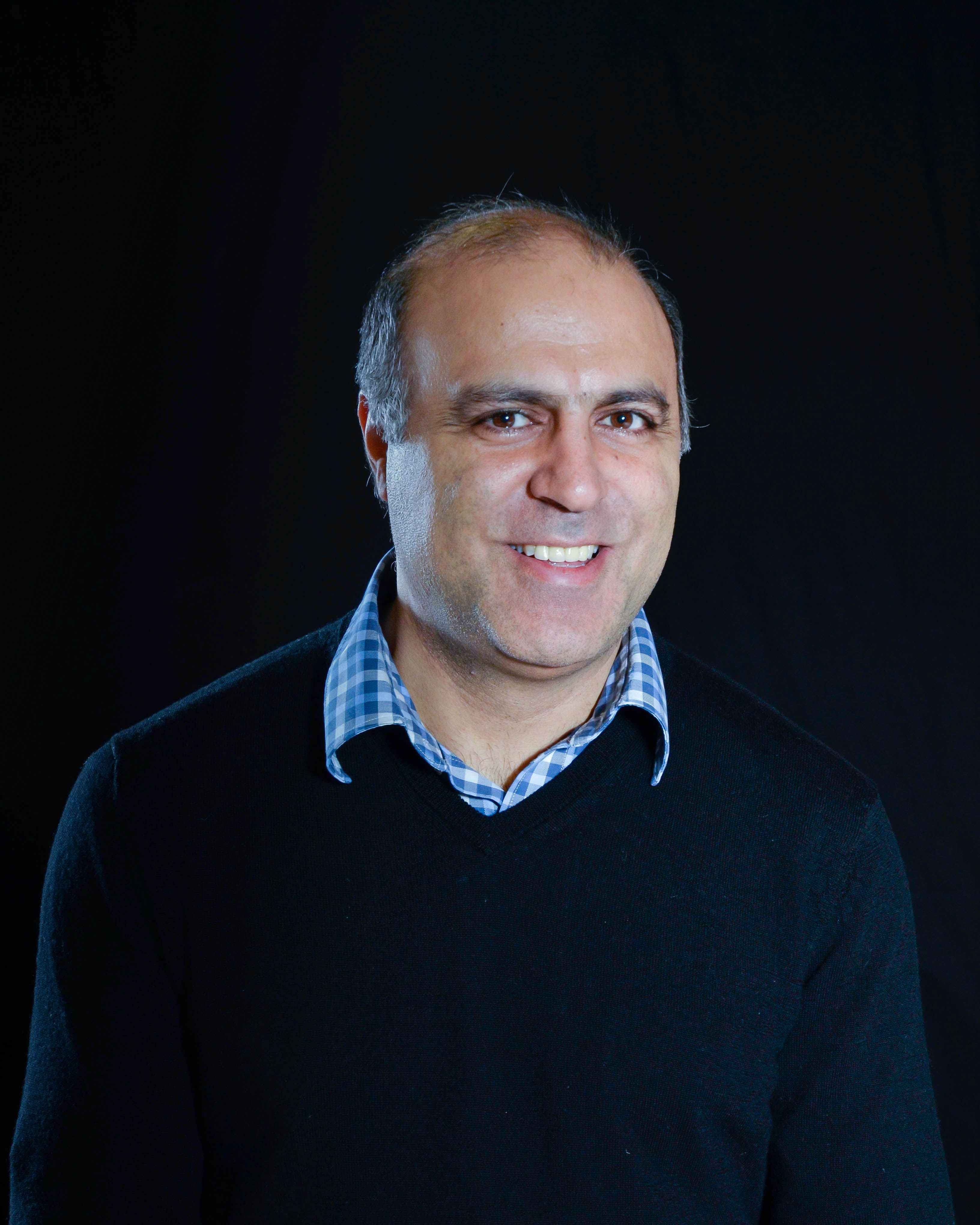}}]{Nasser L. Azad} is an Associate Professor in the Department of Systems Design Engineering at the University of Waterloo and the Director of the Automation and Intelligent Systems (AIS) Group. Before, he was a Postdoctoral Fellow at the University of California, Berkeley. Dr. Azad’s primary research interests lie in (i) intelligent, safe, and secure controls \& automation with applications to automotive systems and autonomous systems like automated vehicles and drones and (ii) innovative applications of AI methods to solve complex modeling, optimization, control, and automation problems.

\end{IEEEbiography}

\end{document}